%% file: main.tex
\documentclass{article}

   \usepackage[nonatbib]{neurips_2022}

\usepackage[utf8]{inputenc} 
\usepackage[T1]{fontenc}    
\usepackage{hyperref}       
\usepackage{url}            
\usepackage{booktabs}       
\usepackage{amsfonts}       
\usepackage{amsmath, amsthm}
\usepackage{mathtools}
\usepackage{nicefrac}       
\usepackage{microtype}      
\usepackage{xcolor}         
\usepackage{float}         
\usepackage{multirow}
\usepackage{wrapfig}

\newtheorem{theorem}{Theorem}

\newcommand{\Ex}[1]{\ensuremath{\mathbb{E}\left[#1\right]}}
\usepackage{wrapfig}
\usepackage{caption}

\title{Explicit Tradeoffs between \\ Adversarial and Natural Distributional Robustness}

\author{%
  Mazda Moayeri \\
  \texttt{mmoayeri@umd.edu} \\
   \And
   Kiarash Banihashem \\
   \texttt{kiarash@umd.edu} \\ 
   \And
   Soheil Feizi \\
   \texttt{sfeizi@cs.umd.edu} \\
   \AND \vspace{-0.5cm}
   \\ 
  Department of Computer Science \\ 
  University of Maryland \\ 
}

\newcommand{\mN}{N}
\newcommand{\opttheta}{{\theta^{\text{opt}}}}
\newcommand{\vecdot}[2]{{\left<#1, #2\right>}}
\newcommand{\R}{{\mathbb{R}}}
\newcommand{\norm}[1]{{\left\| #1\right\|}}
\newcommand{\abs}[1]{{\left| #1\right|}}
\newcommand{\parant}[1]{{\left( #1 \right)}}
\newcommand{\nfs}{{\textnormal{NFS}}}

\usepackage{caption}
\usepackage{subcaption}
\usepackage{todonotes}

\begin{document}

\maketitle

\begin{abstract}

\looseness=-1
Several existing works study either adversarial or natural distributional robustness of deep neural networks separately. In practice, however, models need to enjoy both types of robustness to ensure reliability. In this work, we bridge this gap and show that in fact, {\it explicit tradeoffs} exist between adversarial and natural distributional robustness. We first consider a simple linear regression setting on Gaussian data with disjoint sets of \emph{core} and \emph{spurious} features. In this setting, through theoretical and empirical analysis, we show that (i) adversarial training with $\ell_1$ and $\ell_2$ norms increases the model reliance on spurious features; (ii) For $\ell_\infty$ adversarial training, spurious reliance only occurs when the scale of the spurious features is larger than that of the core features; (iii) 
adversarial training can have {\it an unintended consequence} in reducing distributional robustness, specifically when spurious correlations are changed in the new test domain. Next, we present extensive empirical evidence, using a test suite of twenty adversarially trained models evaluated on five benchmark datasets (ObjectNet, RIVAL10, Salient ImageNet-1M, ImageNet-9, Waterbirds), that adversarially trained classifiers rely on backgrounds more than their standardly trained counterparts, validating our theoretical results. We also show that spurious correlations in training data (when preserved in the test domain) can {\it improve} adversarial robustness, revealing that previous claims that adversarial vulnerability is rooted in spurious correlations are incomplete.

\end{abstract}
\section{Introduction}

Despite continuously improving upon state of the art accuracy on various benchmarks, deep image classifiers remain brittle to distribution shifts, suffering massive performance drops when evaluated on non-i.i.d. data. For example, the accuracy of object detectors trained on ImageNet \cite{imagenet} reduces by 40-45\% on ObjectNet \cite{ObjectNet}, where images are taken within households at various viewpoints and rotations. The reliance of deep models on {\it spurious features}, which correlate with class labels but are irrelevant to the true labeling function \cite{dfr}, is one cause of poor model robustness, as performance degrades when spurious correlations are broken. Indeed, model reliance on spurious features like texture \cite{geirhos} and background \cite{noise_or_signal, cct} is well documented. Of greater concern, deep models in safety critical applications such as detection of pneumonia \cite{zech} and COVID-19 \cite{degrave} have been observed to rely on hospital specific spurious markers, causing poor generalization to new hospitals. 

\begin{figure}
    \centering
    \begin{minipage}{0.37\linewidth}
    \centering
    \includegraphics[width=0.98\linewidth]{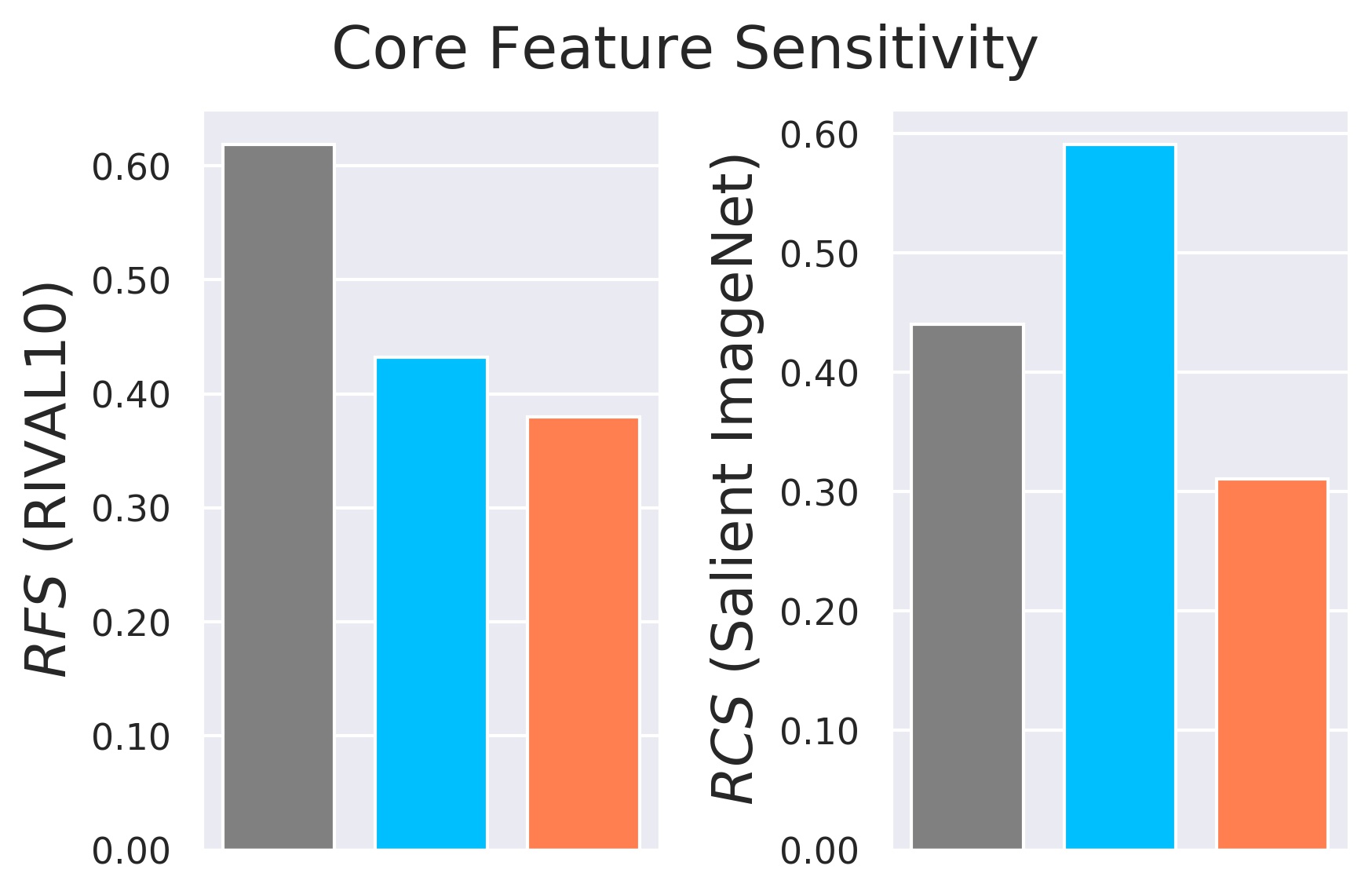}
    \end{minipage}
    \begin{minipage}{0.37\linewidth}
    \centering
    \includegraphics[width=0.98\linewidth]{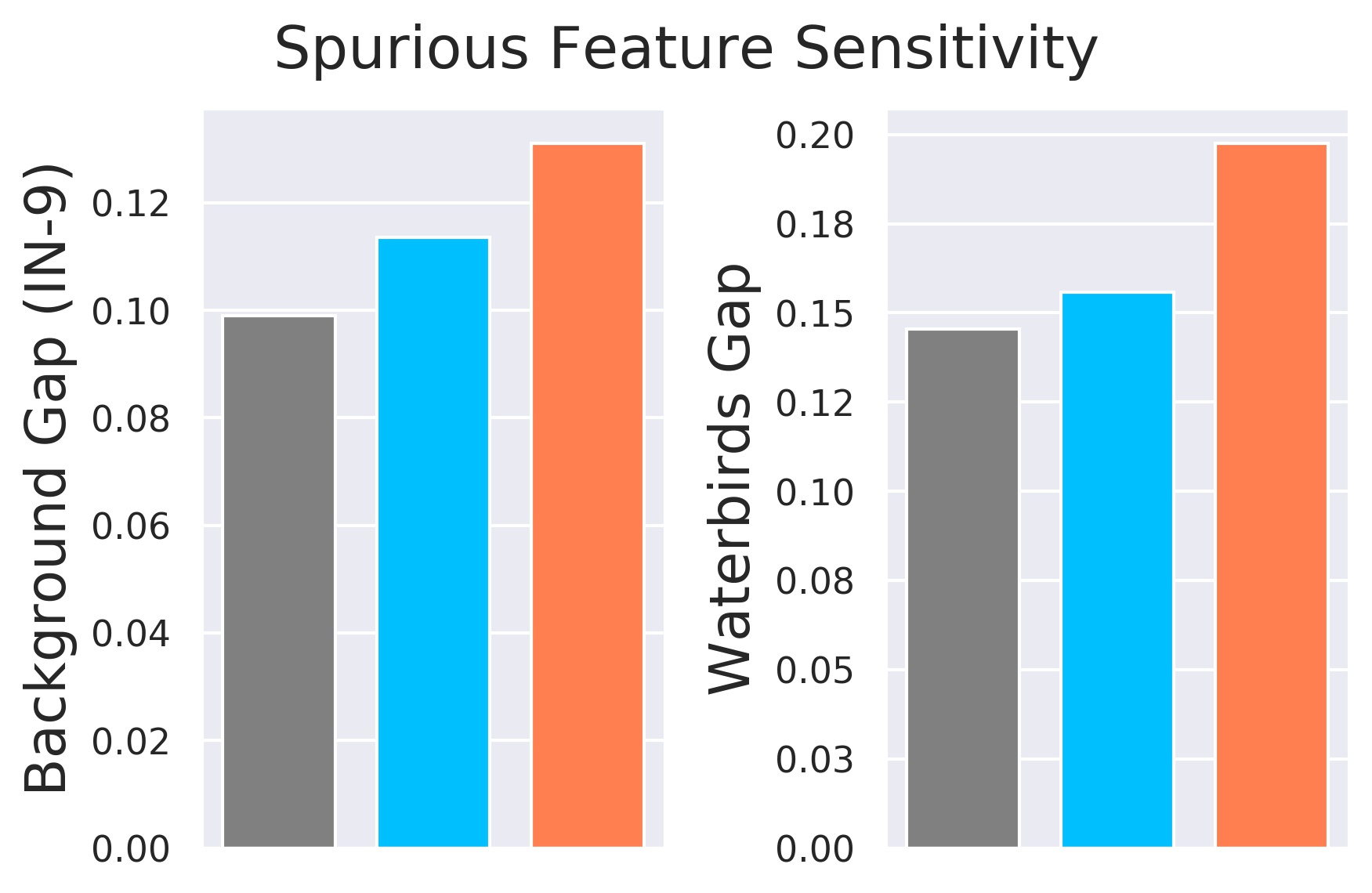}
    \end{minipage}
    \begin{minipage}{0.218\linewidth}
    \centering
    \includegraphics[width=0.99\linewidth]{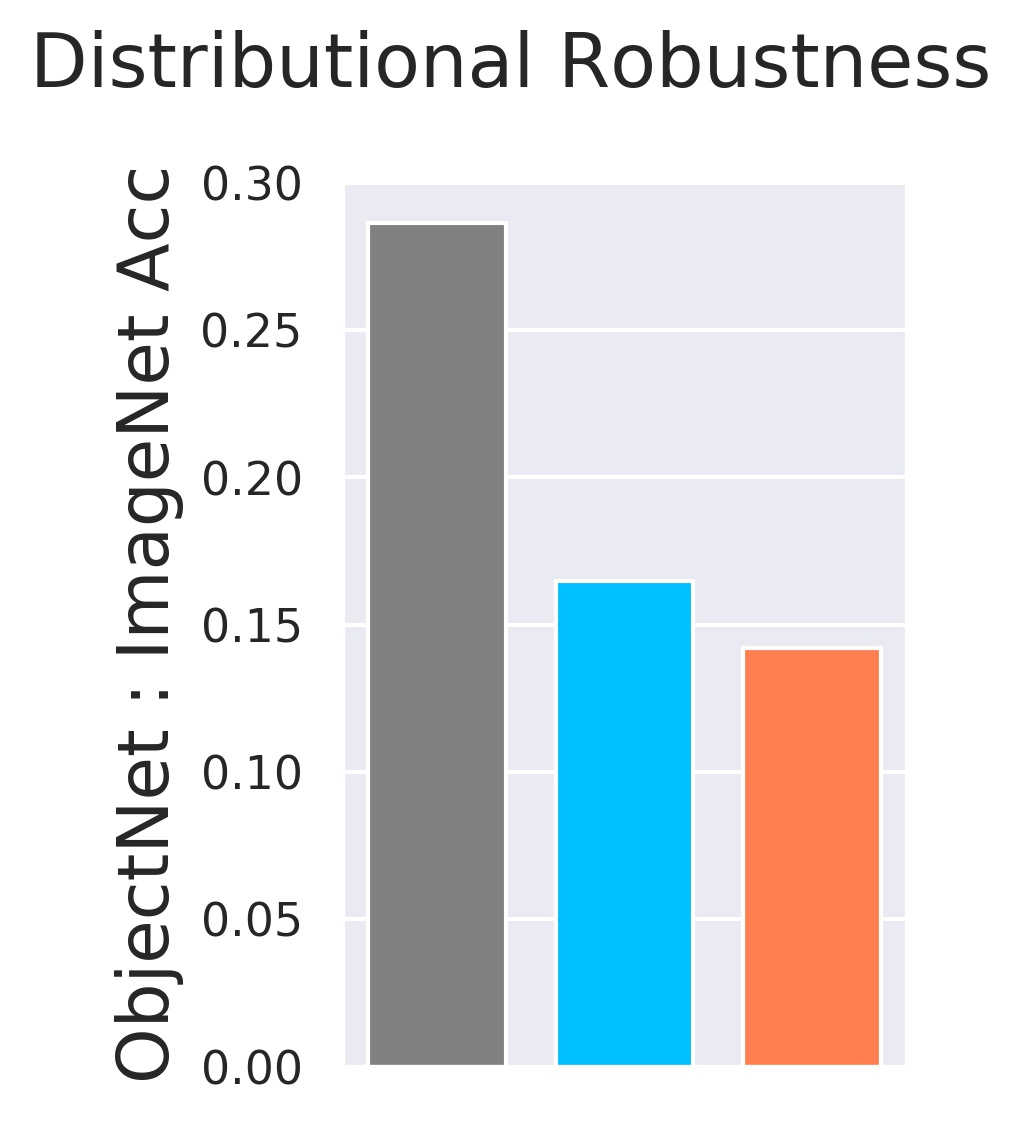}
    \end{minipage}
    \\ 
    \begin{minipage}{0.65\linewidth}
    \centering
    \includegraphics[width=0.7\linewidth]{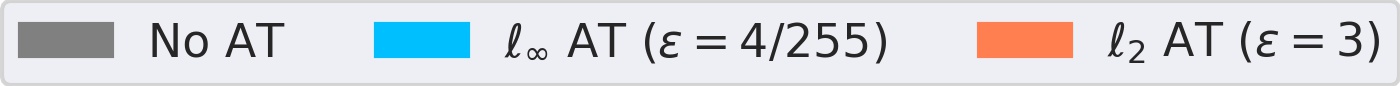}
    \end{minipage}
    \caption{Snapshot of empirical evidence using {\it RIVAL10, Salient ImageNet-1M, ImageNet-9, Waterbirds,} and {\it ObjectNet} benchmarks. Results averaged over ResNet18 and ResNet50. {\bf Adversarial training, especially under $\ell_2$ norm, reduces (increases) sensitivity to core (spurious) features. The increased reliance on spurious features leads to worse distributional robustness}.}
    \label{fig:teaser}
\end{figure}

Adversarial examples \cite{szegedy, goodfellow} pose another troubling distribution shift, where imperceptible input perturbations can cause model accuracy to drop to zero. Many works have been proposed to improve the adversarial robustness of deep models \cite{pat, trades, salman_prov, sahil_secondorder, alex, cohen, fast_at, free_at}, including the widely popular {\it adversarial training}, where inputs are augmented via adversarial attack during training \cite{pgd}. While spurious correlation robustness has also attracted lots of attention \cite{irm, dfr, jtt, dro}, the two problems are most often considered independently, despite both being essential to the safe and reliable deployment of deep models in the wild.

In the few works that do consider adversarial and spurious correlation robustness in tandem, the prevailing argument is that the origin of adversarial vulnerability is that the model focuses on spurious correlations that can be manipulated \cite{causaladv, cama, ftrs_not_bugs}. However, recently, noise-based analyses on RIVAL10 \cite{rival10} and Salient ImageNet-1M \cite{salientImageNet1M} datasets suggest adversarial training may actually increase model sensitivity to spurious features; a result that is both counter-intuitive and in direct contrast to existing ideology. 

To better understand this observation, we first appeal to a simple linear regression problem on Gaussian data with disentangled {\it core} and {\it spurious} features. In this setting, we theoretically show 
\begin{itemize}
    \item Adversarial training under $\ell_2$ and $\ell_1$ norms increase model reliance on spurious features, as using spurious features forces an attacker to spread its budget over additional features.
    \looseness=-1
    \item For $\ell_\infty$ adversarial training, increased spurious feature reliance only occurs when {\it the scale} of the spurious feature is larger than that of the core features. That is, spurious features are used when perturbations that corrupt core features are too small to disrupt spurious correlations. 
    \item Due to increased spurious feature reliance, there is an {\it explicit tradeoff} between adversarial and distributional robustness. Specifically, we show that adversarial training decreases model robustness to distribution shifts in the test domain where spurious correlations are broken.  
\end{itemize}
\looseness=-1
To validate our theory, we evaluate twenty models adversarially trained using $\ell_2$ and $\ell_\infty$ projected gradient descent \cite{robust_models_transfer}. Specifically, we inspect performance on multiple spurious robustness benchmarks over synthetic (ImageNet-9 \cite{noise_or_signal}, Waterbirds \cite{dro}) and real (RIVAL10 \cite{rival10}, Salient ImageNet-1M \cite{salientImageNet1M}, ObjectNet \cite{ObjectNet}) datasets. Figure \ref{fig:teaser} summarizes our experiments, where we find that adversarially trained models consistently show greater sensitivity to spurious features compared to standardly trained baselines, with the effect more dramatic for $\ell_2$ adversarial training than $\ell_\infty$\footnote{High $RCS$ for $\ell_\infty$ AT models is due to reduced {\it scale} of contextual bias in Salient ImageNet since the data diversity weakens background correlations. In $\ell_1$ and $\ell_2$ adversarial training, models rely on spurious features regardless of their scale. See details in Section \ref{sec:theory}.}. Finally, we show that the presence of spurious correlations in training data (when preserved in test domain) can improve adversarial robustness, with stronger spurious correlations leading to greater accuracy under attack.

Our work combines two prevalent but often separately considered notions of robustness, yielding surprising theoretically-derived and empirically-supported results. We hope our contributions grant insight to both adversarial and distributional robustness communities, and emphasize the need for holisitic evaluations of model robustness.

\section{Review of Literature}

\noindent{\bf Adversarial Robustness.} Since adversarial examples were first observed in deep models \cite{szegedy, goodfellow}, the phenomenon has been extensively studied. New attacks \cite{cw_attack, recolor, deepfool, spatial} and defenses \cite{pat, distill, simcat, dmat} are introduced frequently, in a game of cat and mouse where the attacker generally has the upper hand \cite{athalye}. Certified defenses seek to break this cycle by offering provable robustness guarantees \cite{sahil_secondorder, alex, salman_prov, cohen}. Arguably the most popular defense is \emph{adversarial training} \cite{pgd} where images are augmented with adversarial perturbations during training, amounting to a min max optimization. 

\looseness=-1
\noindent{\bf Natural Distributional Robustness.} In contrast to synthetic adversarial perturbations, many works seek to characterize the robustness of deep models to naturally occurring distribution shifts, for instance due to common corruptions (noise, blur, etc) \cite{imagenet_c} or changes in rendition \cite{manyfaces, sketch}. \cite{wilds} compiles ten benchmarks of realistic distribution shifts over diverse applications (medical, economic, etc). Many algorithms have been proposed to improve out-of-distribution robustness \cite{jtt, dro, irm, vrex}, though in comprehensive evaluations, their gains over empirical risk minimization are marginal, as they often only hold for {\it certain} distribution shifts \cite{ood_bench, manyfaces}. \cite{ood_bench} identifies {\it diversity} and {\it correlation} shifts as two key dimensions to OOD robustness; our work focuses on the latter.   

\looseness=-1
\noindent{\bf Spurious Correlations.} Solely optimizing for accuracy leads deep models to rely on {\it any} patterns predictive of class in the training domain. This includes {\it spurious} features, which are irrelevant to the true labeling function. Natural image datasets are riddled with spurious features \cite{distractors_kolesnikov, iclr_salientImageNet}. Spurious feature reliance becomes problematic under distribution shifts that break their correlation to class labels: sidewalk segmentation struggles in the absence of cars \cite{sidewalk}, familiar objects cannot be recognized in unfamiliar poses \cite{adv_pose} or uncommon settings \cite{focus, ObjectNet}, etc. A natural and ubiquitous spurious correlation in vision is image backgrounds, observed in numerous prior works to be leveraged by models for classification \cite{noise_or_signal, rival10} and object detection \cite{elephant}. Spurious correlations also relate to algorithmic biases \cite{si_score, geirhos}, with implications for fairness \cite{fairness, gender, recidivism, khani}, reflecting the importance of this issue.

\looseness=-1
Accordingly, many works seek to improve spurious correlation robustness. Families of approaches include optimizing for worst group accuracy \cite{dro, dro_like, dro_like2, dro_like3, jtt}, learning invariant latent spaces \cite{hassani, irm}, appealing to meta-learning \cite{meta} or causality \cite{causal1, causal2}. Our work does not focus on mitigation methods, but instead sheds insight on how optimizing for a different notion of robustness (i.e. adversarial) affects spurious feature reliance, and consequently, natural distributional robustness. Generally, models trained under ERM are believed to have a propensity to use spurious features, especially when they are easy to learn, due to bias of learners (algorithmic and human) to absorb simple features first \cite{simplicity_bias} and take shortcuts \cite{shortcuts}. However, recent work suggests that core features are still learned under ERM even when spurious features are favored, and simple finetuning on data without the spurious feature can efficiently reduce spurious feature reliance without full model retraining \cite{dfr}. 

\noindent{\bf Unintended Outcomes of Adversarial Training.} Adversarial training achieves improved accuracy under attack, but comes at the cost of standard accuracy, with multiple works provably demonstrating this tradeoff \cite{trades, Dobriban2020ProvableTI}. Notably, \cite{javanmard2020precise} inspires our theoretical analysis, though we focus on the effect of adversarial training on out-of-distribution (OOD) robustness to spurious correlation shifts, rather than its effect on standard accuracy. A more positive outcome is that adversarial training leads to perceptually aligned gradients \cite{pag}, with applications to model debugging \cite{debug_sahil, debuggable}, and further, transfer learning on adversarially robust features yields better accuracy on downstream tasks compared to features learned from standard training \cite{robust_models_transfer}, despite having lower accuracy on the original task. 

To our knowledge, robustness to adversarial and natural distribution shifts have not been studied in tandem. However, spurious correlations are at times mentioned with adversarial robustness, usually in claims that the origin of adversarial vulnerability is in model's focus on (imperceptible) spurious features \cite{causaladv, cama, ftrs_not_bugs}. Our results create tension with the contrapositive of their argument, as we show that mitigating adversarial vulnerability (via adversarial training) results in {\it increased} spurious feature reliance. We do this analytically in a simple linear regression setting (Section \ref{sec:theory}), and empirically on multiple benchmarks, with an emphasis on natural spurious features (i.e., backgrounds) in our experiments (Section \ref{sec:experiments}). Further, we even demonstrate a case where the presence of a spurious feature leads to {\it improved} adversarial robustness (Section \ref{sec:reverse})). We note that the spurious features we observe to be positively associated with adversarial robustness may be distinct from those that prior works claim contribute to adversarial vulnerability. However, our result of adversarial training leading to increased spurious feature reliance (of any kind) is novel and contrary to common understanding. Given the critical nature of adversarial and spurious correlation robustness for model security, reliability, and fairness, the significance of our result in revealing potential misconceptions on the interplay of these two crucial modes of robustness should not be understated.

\cite{rival10} and \cite{salientImageNet1M} recently observed decreased core sensitivity on a handful of $\ell_2$ adversarially trained models, in spirit with our findings, but with no explanation. We offer the first rigorous analysis of this counterintuitive phenomenon, evaluating $5$ to $10$ times as many models in $5$ times as many settings. More importantly, we theoretically prove that adversarial training increases spurious feature reliance, contributing novel fundamental insight as to how optimizing for adversarial robustness can lead to reduced robustness to natural distribution shits, uncovering important effects like the norm of adversarial training and the scale of spurious features at play. 


\section{Theoretical Analysis on Linear models}
\label{sec:theory}

We begin by analysing the effects of adversarial training on a simple linear regression model.
Consider the model $Y=\vecdot{X}{\opttheta} + W$ where
$X \in \R^m$ is the input variable, $\opttheta \in \R^m$ is the optimal parameter,
$\vecdot{.}{.}$ represents the inner product
and $W \in \R$ is a noise variable.
We assume that the input variables follow a multivariate Gaussian distribution
$\mN(0, \Sigma)$ where $\Sigma \in \R^{m\times m}$ is the covariance matrix
and further assume that $W$ is sampled from the Gaussian distribution $\mN(0, \sigma_{w}^2)$.
~\\
We assume that the set of features $[m]$ consists of two groups,
the \emph{core features} $C$ and the \emph{spurious features} $S$.
Without loss of generality, we assume that $C = \{1, \dots, p\}$ and
$S = \{p + 1, \dots, m\}$. 
We assume that the optimal parameter $\opttheta$ has non-zero
entries on the set $C$ only.
This implies that the output depends on the input only through the
core features
and conditioned on the core features, it is independent of the spurious ones.
More formally, we assume that
$  Y \perp X_{S} | X_{C}$
where $X_{S}$ and $X_{C}$ represent the core and spurious subsets of the input, respectively.

For loss functions, we define the
\emph{standard loss} function
 as $
  L(\theta) = \Ex{(y - \vecdot{X}{\theta})^2},
  $,
and the \emph{adversarial loss}
 function as
\begin{align}
  L_{p, \epsilon}(\theta) = \Ex{\max_{\norm{\delta}_p \le \epsilon} (Y - \vecdot{X + \delta}{\theta})^2},
  \label{eq:loss_adv}
\end{align}
where $p$ is the attack norm and $\epsilon$ represents the norm budget.

We first show an equivalent form of \eqref{eq:loss_adv} that is more amenable to analysis.
\begin{theorem}\label{thm:only}
  Assume that $Y=\vecdot{X}{\opttheta} + W$ where
  $W \sim \mN(0, \sigma_{w}^2)$ is independent of $X$ and
  $\opttheta \in \R^m$ is a fixed parameter.
  Assume further that $X$ follows the distribution $\mN(0, \Sigma)$
  and
  define $\sigma_\theta^2$ as 
  $(\theta - \opttheta)^T\Sigma(\theta - \opttheta) + \sigma_w^2$.
  The loss function \eqref{eq:loss_adv} is equivalent to
  \begin{align}
    L_{p, \epsilon} (\theta) = c_2 \cdot \sigma_{\theta}^2 + (c_1\sigma_{\theta} + \epsilon \cdot \norm{\theta}_q)^2
    \label{eq:thm_statement}
  \end{align}
  where $c_1 = \sqrt{\frac{2}{\pi}} < 1$, $c_2 = 1 - c_1^2$ and $\norm{.}_q$ is the dual norm of
  $\norm{.}_p$, i.e, $\frac{1}{p} + \frac{1}{q} = 1$.
  Furthermore, the above formulation is convex in $\theta$.
\end{theorem}
The above result is similar to Proposition 3.2 in \cite{javanmard2020precise} which provides characterization results for the $\ell_2$ norm. The key distinction of our results is providing a simple convex formulation of the robust minimization problem, allowing the results to be easily generalized for an arbitrary $\ell_p$ norm. 
The
theorem shows that the optimal value
$\widehat{\theta}$ minimizing the adversarial losss $L_{p, \epsilon}(\theta)$ 
\emph{is not} $\opttheta$ and in general,
may be non-zero on the set of spurious features $S$.
This means that adversarial training directs the model
towards using the spurious correlations in order to increase robustness.

The proof of the theorem is provided in the Appendix. The main structure of the proof is similar to that of \cite{javanmard2020precise}. We first show that the inner maximization problem of \eqref{eq:loss_adv} can be rewritten as
\begin{align}
    \max_{\norm{\delta} \le \epsilon} (Y - \vecdot{X + \delta}{\theta})^2
     = \left(
       \abs{Y - \vecdot{X}{\theta}} + \epsilon \cdot \norm{\theta}_q
     \right)^2.
     \label{eq:loss_identity_dual_norm}
  \end{align}
This allows us to rewrite \eqref{eq:loss_adv}
as
\begin{align*}
    L_{p, \epsilon} &= \Ex{
      \parant{
        \abs{ Y - \vecdot{X}{\theta}} + \epsilon \cdot \norm{\theta}_q
      }^2
    }
    \\&=
    \Ex{
      \parant{
      Y - \vecdot{X}{\theta}
      }^2
    }
    + \epsilon^2 \cdot \norm{\theta}_q^2  + 2\cdot \epsilon \cdot \norm{\theta}_q \cdot \Ex{\abs{Y - \vecdot{X}{\theta}}}
    \\&\overset{(a)}{=}
    \Ex{
      \parant{
        \vecdot{X}{\theta - \opttheta}
        + W
      }^2
    }
    + \epsilon^2 \cdot \norm{\theta}_q^2  + 2\cdot \epsilon \cdot
    \norm{\theta}_q \cdot \Ex{\abs{\vecdot{X}{\theta - \opttheta} + W}},
  \end{align*}
  where for $(a)$ we have used the fact that $Y = \vecdot{X}{\theta} + W$.
  Using the fact that $X$ and $W$ are Gaussian, we show that this is equal to $ \sigma_{\theta}^2 + \epsilon^2 \cdot \norm{\theta}_q^2 + 2 \cdot c_1 \cdot  \epsilon \cdot \norm{\theta}_q \cdot \sigma_{\theta}$
  which we can further simplify to  \eqref{eq:thm_statement} with some algebraic manipulation. 
  Next, using standard  techniques for analysing composition of convex functions, we show that $\sigma_{\theta}^2$ and 
  $ (c_1\sigma_{\theta} + \epsilon \cdot \norm{\theta}_q)^2$ are both convex in $\theta$, implying that
  $L_{p, \epsilon}(\theta)$ is convex in $\theta$ as well.

Using the convex formulation in Theorem \ref{thm:only}, we evaluate the 
linear model
for different values of parameters to better understand the reliance of the model on spurious features.
In our experiments, we consider a simple model with 5 features, the first two of which are core features.~\\
We let $\eta$ be a parameter controlling 
the correlation degree between the core and spurious features, with larger values corresponding to higher correlation.
For the distribution of the data, define the matrix $\widetilde{Q}$ and $Q$ matrices as
\begin{align}
  \widetilde{Q} = 
  \begin{bmatrix}
    1 & \frac{1}{2} & 0 & 0 & 0\\
    \frac{1}{2} & 1 & 0 & 0 & 0\\
    \eta & \eta & 1 & 0 & 0\\
    \eta & \eta & 0 & 1 & 0\\
    \eta & \eta & 0 & 0 & 1\\
  \end{bmatrix},\quad
  Q_{i, j} = \frac{\widetilde{Q}_{i, j}}{\sqrt{\sum_{i, j'} \widetilde{Q}_{i, j'}^2}}
  \label{eq:def_q}
\end{align}
Note that $Q$ is obtained by normalizing the rows of $\widetilde{Q}$.
Each row of the $Q$ matrix corresponds to an input feautre.
We let $\Sigma$ take the value $QQ^T$. This is equivalent to sampling $X$ from the distribution $Q \mathcal{N}(0, I)$. Throughout our experiments, we set $\sigma_{w}=0.1$.

\begin{figure}[ht]
    \centering
	\begin{subfigure}{0.3\textwidth}
    	\centering
		\resizebox{0.98\linewidth}{!}{\includegraphics{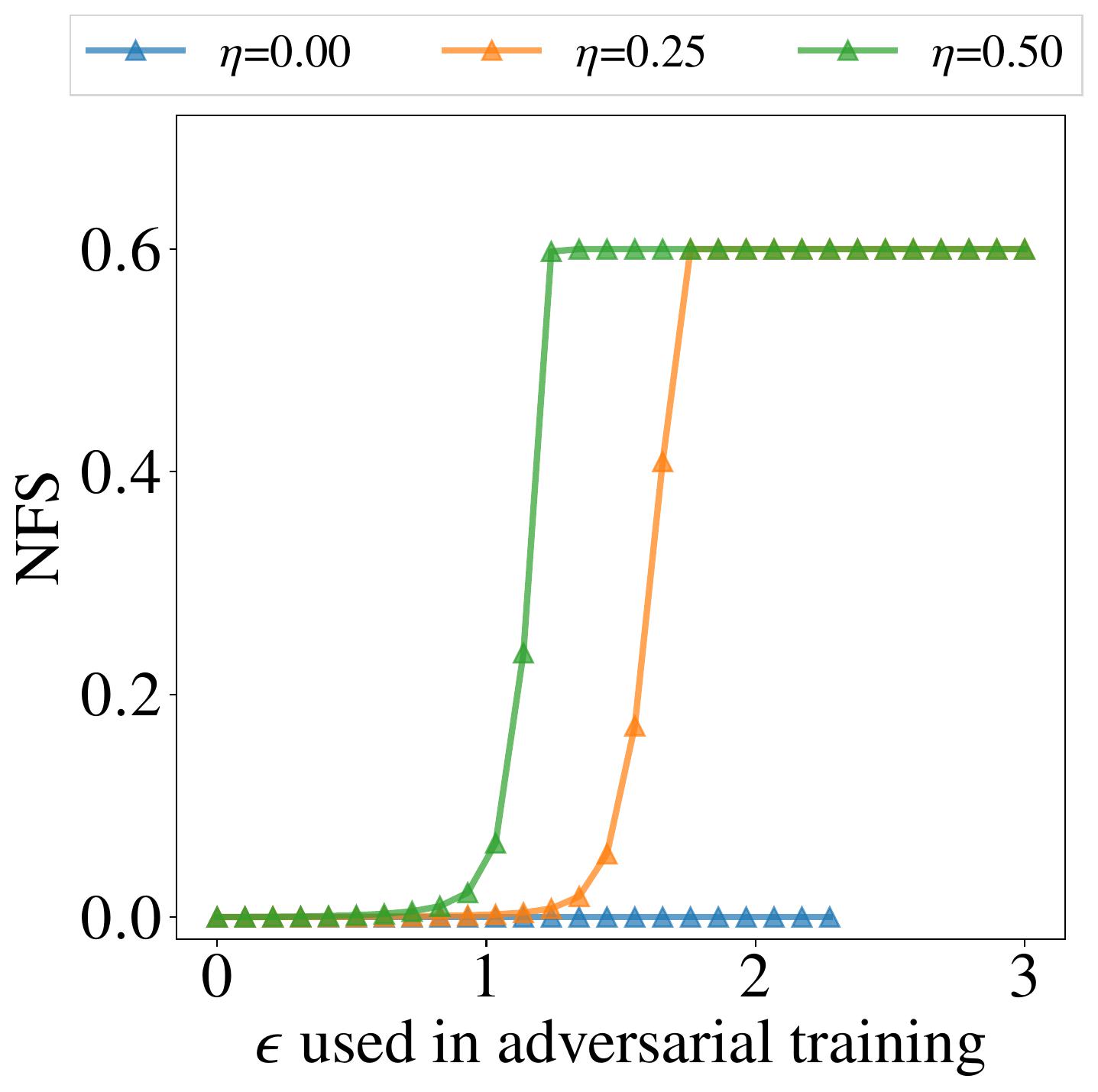}}
		\caption{$\ell_1$ norm}
	\end{subfigure}
    \begin{subfigure}{0.3\textwidth}
    	\centering
		\resizebox{0.98\linewidth}{!}{\includegraphics{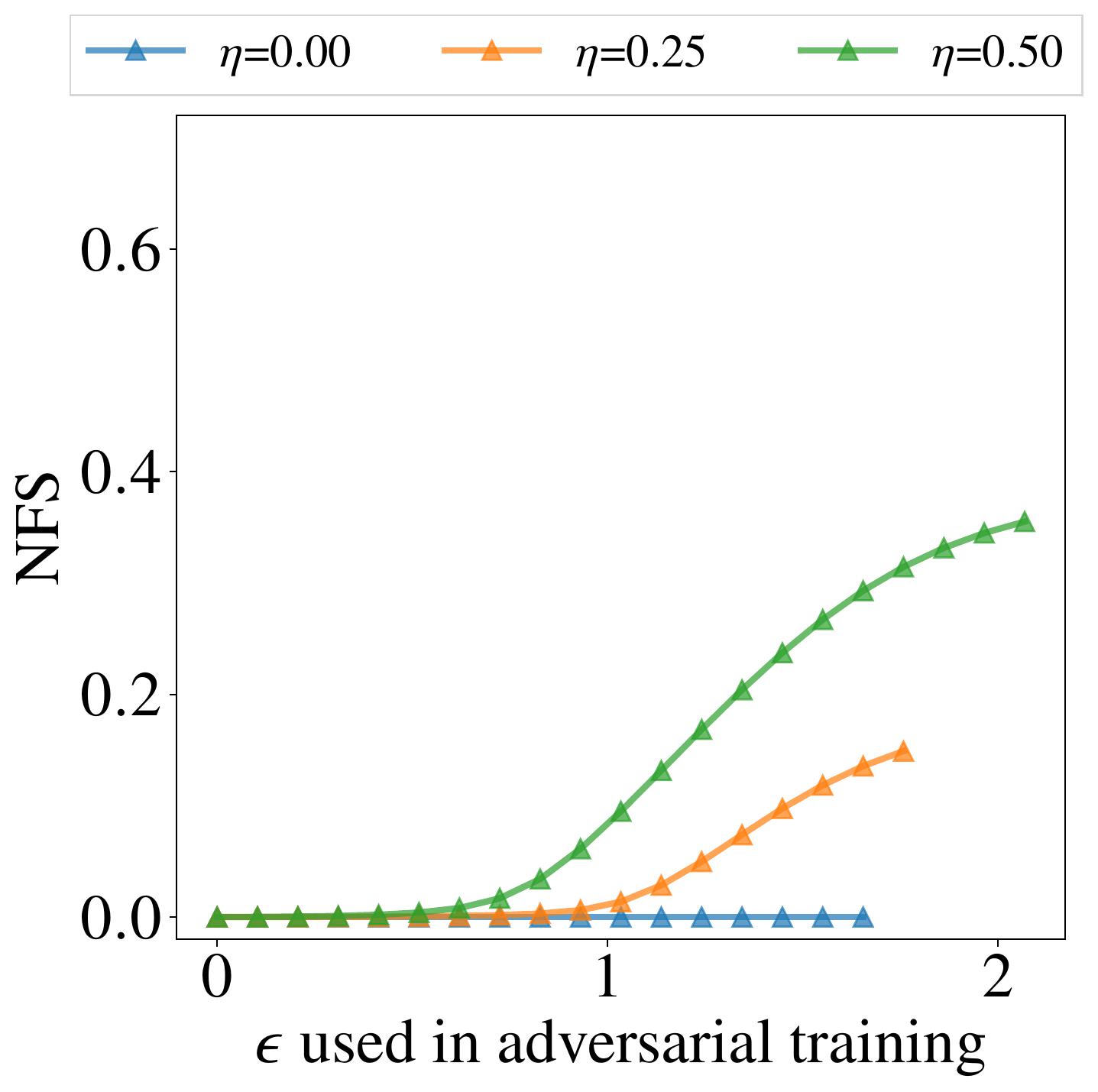}}
		\caption{$\ell_2$ norm}
	\end{subfigure}
	\begin{subfigure}{0.3\textwidth}
    	\centering
		\resizebox{0.98\linewidth}{!}{\includegraphics{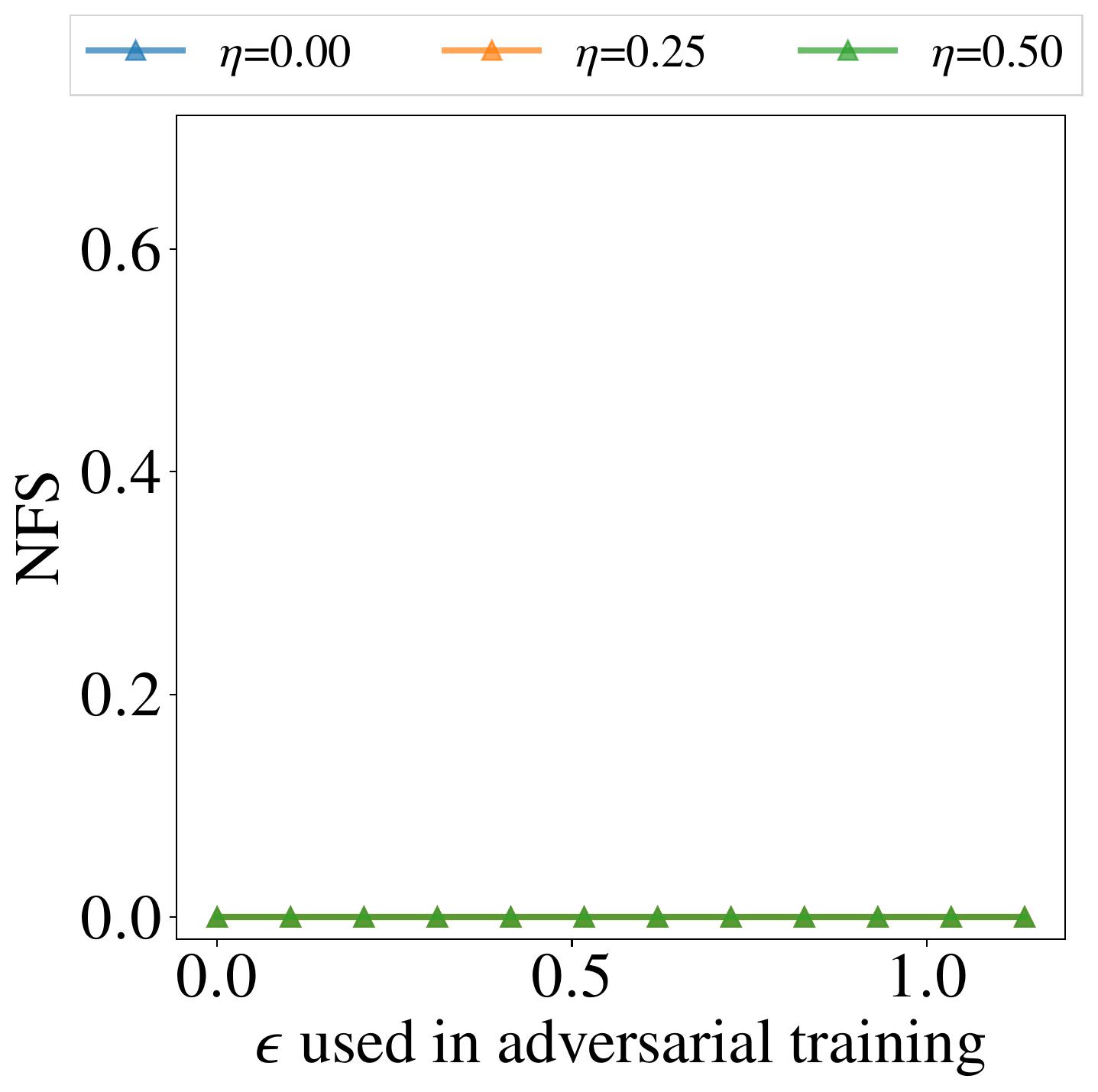}}
		\caption{$\ell_\infty$ norm}
    \label{fig:norm_frac_inf}
	\end{subfigure}
    \caption{Reliance of the adversarially trained model on spurious features as measured by \nfs{} (see Equation \eqref{eq:nfs}) for different choices of
      $\ell_p$ norm and different values of
      adversarial budget
      $\epsilon$ and spurious correlation parameter $\eta$.
    }
    \label{fig:norm_frac}
\end{figure}

For an arbitrary vector $\theta$,
we define its Norm Fraction over Spurious feaures $\nfs(\theta)$ as
\begin{align}
    \nfs(\theta)
    =
    \frac{\sum_{i \in S}\theta_i^2}{\sum_{j}\theta_j^2}
    \label{eq:nfs}
\end{align}
Intuitively, $\nfs$ measures the degree to which a model relies on spurious features.
For our first experiment,
we perform adversarial training with varying parameter $\epsilon$ to obtain a predictor $\widehat{\theta}$ and evaluate its $\nfs$.
We consider different choices of the $\ell_p$ norm as well as the correlation parameter $\eta$
The results can be seen in Figure \ref{fig:norm_frac}.
\vspace{-1em}

As seen in Figure \ref{fig:norm_frac}, for the $\ell_1$ and $\ell_2$ norms, increasing the adversary's budget causes the model to rely more on the spurious features.
To understand why this happens, it is helpful to consider a game theoretic perspective:
If the model only looks at the core features, then the adversary will only need to perturb these features. Thus, even though the spurious features are normally less suitable for prediction, they now have the advantage of being less perturbed.
Assuming $\epsilon$ is large enough, the model will be better off looking at these features in forming its prediction.
Of course, if the model only looks at the spurious features, core features will become even more
informative as they would be unperturbed as well. In the game's equilibrium, the model would use both the core and spurious
features, relying more on the spurious features with increased values of $\epsilon$.

Interestingly, this reasoning does not always apply for the $\ell_{\infty}$ norm. Indeed, if the model were to only look
at the core features, the adversary may still perturb the spurious features with no extra cost.
This is because the $\ell_{\infty}$ norm only measures the \emph{maximum} perturbation in each feature and as long as the perturbations on the spurious features 
are not larger than the perturbations on the core features, the norm would not change.
The results of Figure \ref{fig:norm_frac_inf} 
support this as the adversary does not rely on the spurious features for its predictions.

Importantly, however, in some cases the adversary may require a larger budget to perturb the spurious features compared to
core features.
Indeed, if we were to scale a feature by multiplication with a large number,
then the adversary would require a larger budget to perturb that feature as the budget for that feature has effectively decreased.
We can therefore expect that for the $\ell_{\infty}$ norm, scaling up the spurious features would cause
the model to rely on them for prediction. 
\begin{wrapfigure}[17]{r}{0.35\textwidth}
\vspace{-2em}
\begin{minipage}{0.35\textwidth}
    \begin{figure}[H]
  \centering
  \includegraphics[width=\textwidth]{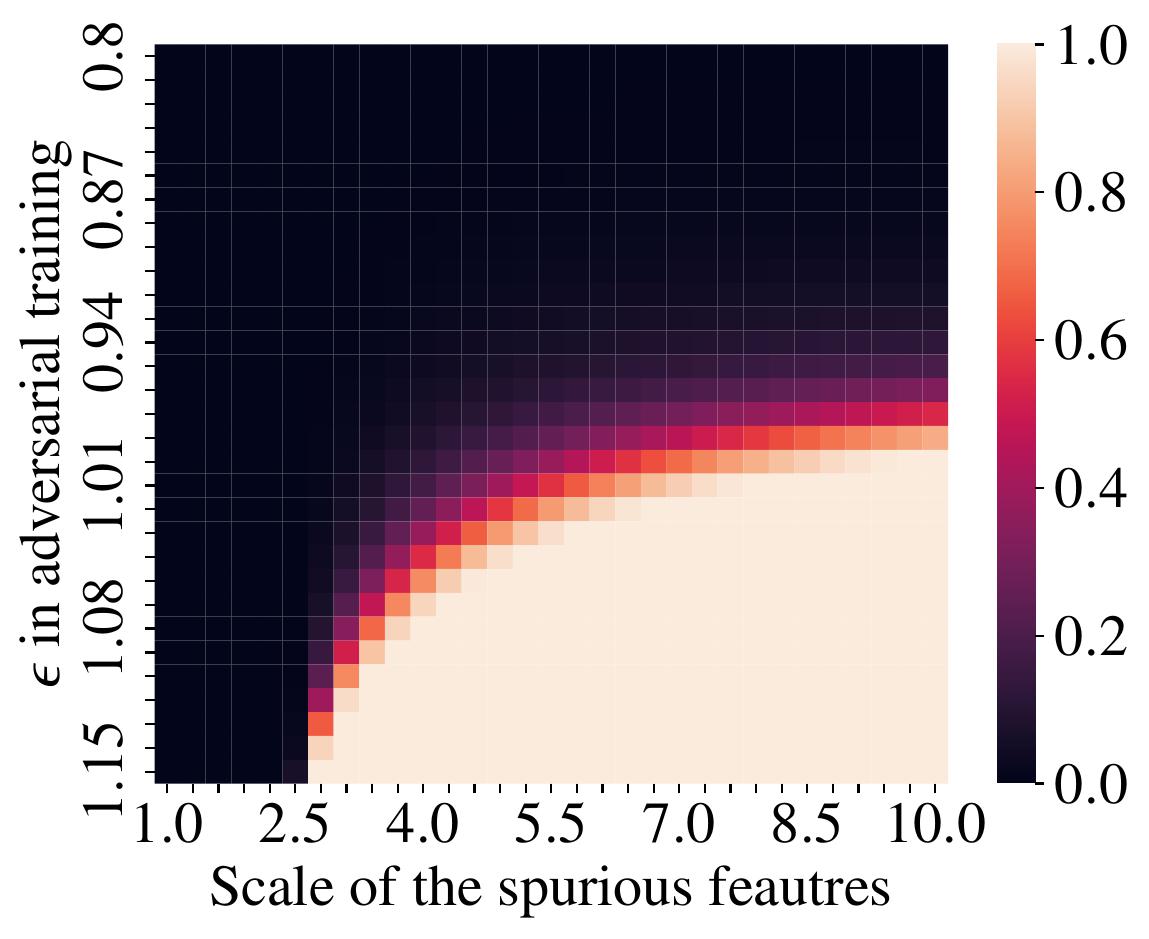}
  \caption{Norm fraction over spurious features (\nfs{}) with an $\ell_{\infty}$ constrained adversary as a function of
    the scale of the spurious features and perturbation budget $\epsilon$.
  }
  \label{fig:heatmap}
\end{figure}
\end{minipage}
\end{wrapfigure}
Figure \ref{fig:heatmap} shows that
this is indeed the case.
The figure shows the norm fraction over spurious (\nfs{}) after scaling the spurious features for different values of $\epsilon$. The scaling is done by multiplying the rows of $Q$ corresponding to the spurious features by a scaling parameter.
As seen in Figure \ref{fig:heatmap}, larger values of the scaling parameter, as well as larger values of $\epsilon$,  cause the model to become more reliant on the spurious features.

\begin{figure}[H]
\vspace{-1 em}
    \centering
	\begin{subfigure}{0.3\textwidth}
    	\centering
		\resizebox{0.98\linewidth}{!}{\includegraphics{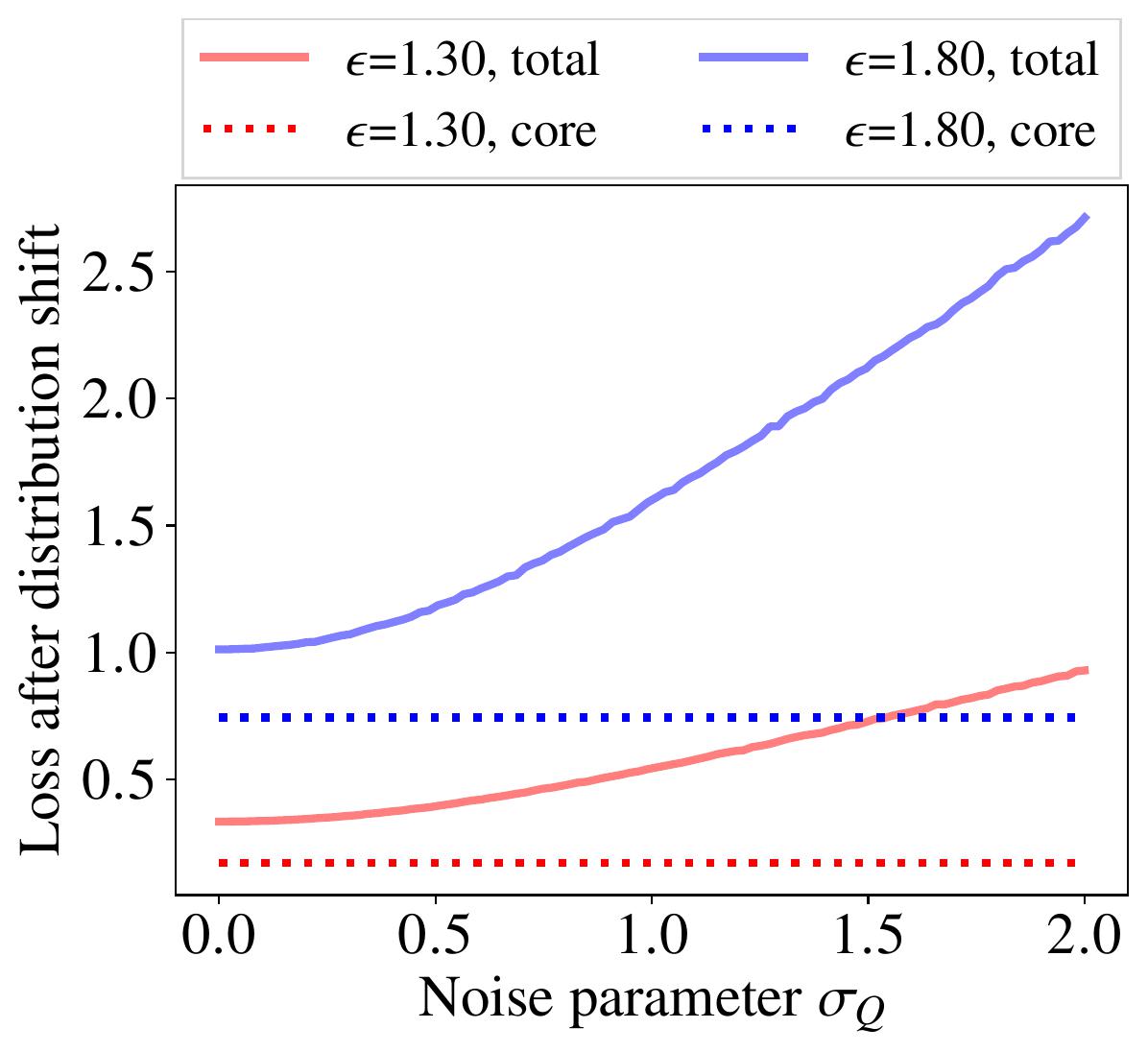}}
		\caption{$\ell_1$ norm}
	\end{subfigure}
	\begin{subfigure}{0.3\textwidth}
    	\centering
		\resizebox{0.98\linewidth}{!}{\includegraphics{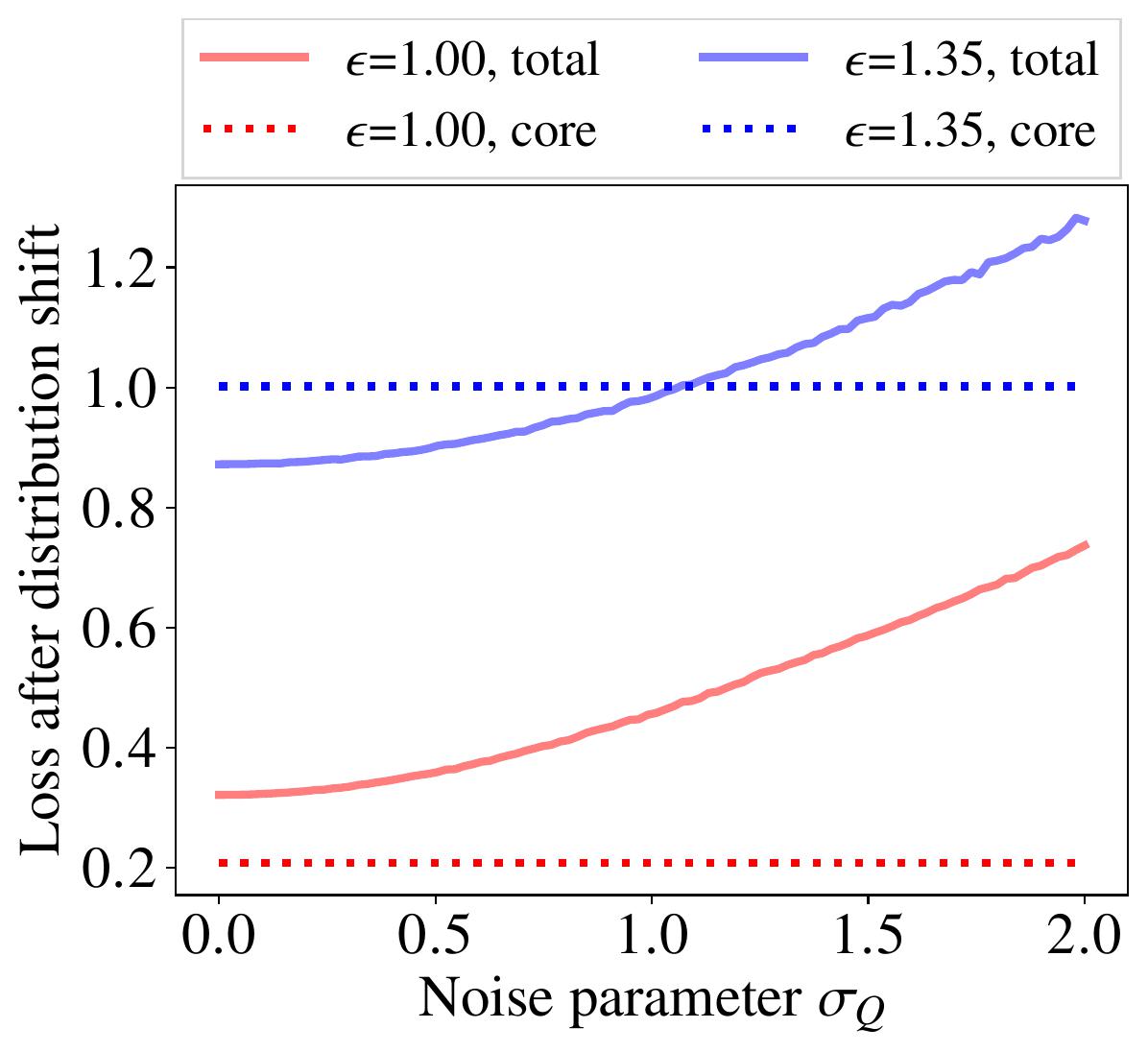}}
		\caption{$\ell_2$ norm}
	\end{subfigure}
	\begin{subfigure}{0.3\textwidth}
    	\centering
		\resizebox{0.98\linewidth}{!}{\includegraphics{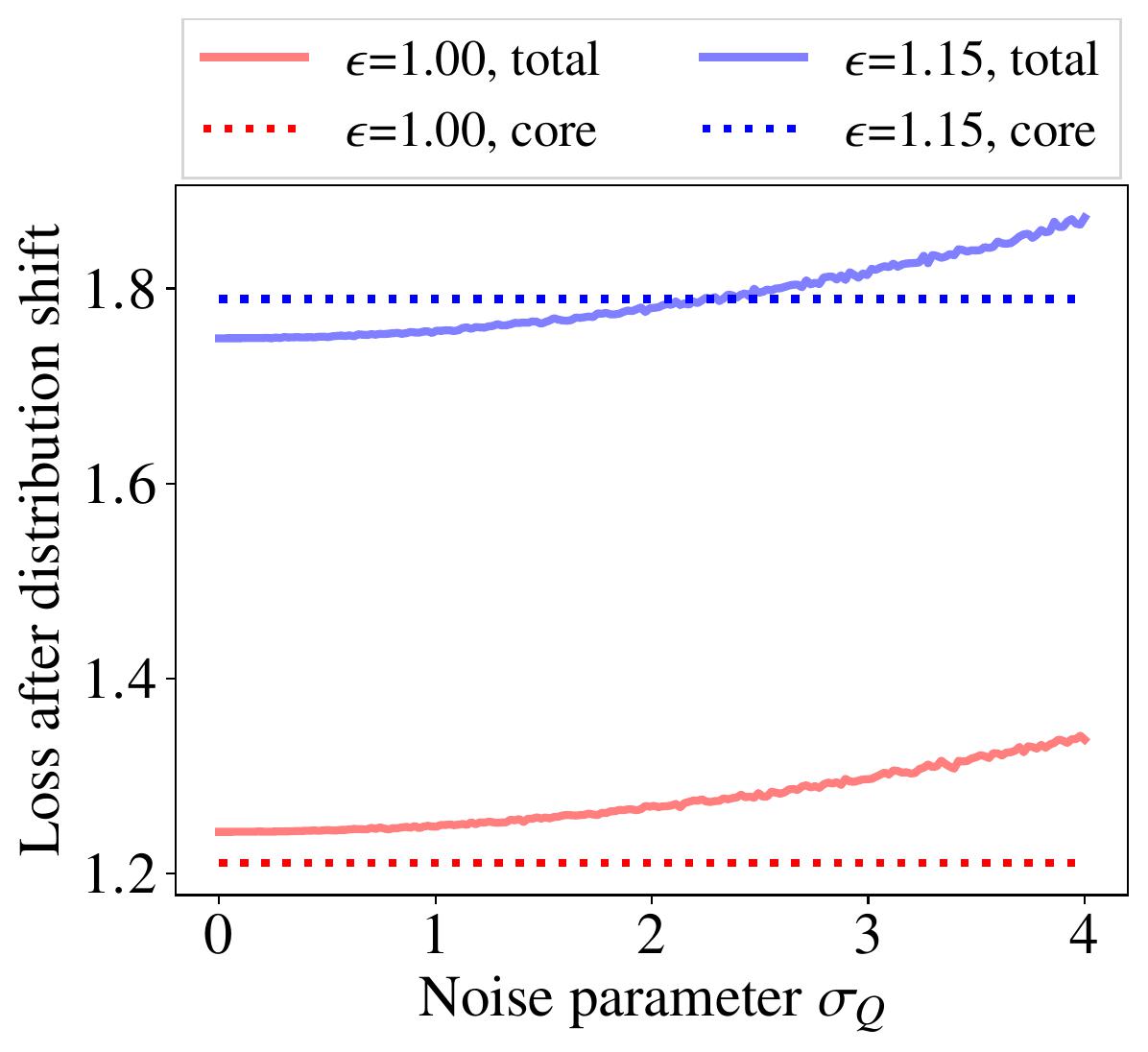}}
		\caption{$\ell_{\infty}$ norm}
	\end{subfigure}
	\caption{Effect of reliance on spurious features on distributional robustness.
	Each figures compares two models, one using only the core features, and another using all of the features (denoted by ``core'' and ``total'' respectively). For the $\ell_1$ and $\ell_2$ norms, the scale of the spurious features is $1$ while for the $\ell_{\infty}$ norm, the scale is set to 3.
	}
    \label{fig:noise}
\end{figure}

Next, we measure the effect of using spurious features on the model's distributional robustness.
To do this, we train two adversarial models.
The first model, which we denote by ``core'', uses only the core features while
the second model, denoted by ``total'', uses all of the features.
We then simulate a distribution shift that breaks the spurious correlations by adding random 
Gaussian noise to the spurious rows of matrix $Q$ defined in \eqref{eq:def_q}.
Specifically, for each entry of $Q$ in a spurious row, we add a noise sampled from $\mN(0, \sigma_{Q}^2)$ where $\sigma_Q$ is a noise parameter.
We use a fixed value of $\eta=0.25$ and vary the parameter $\epsilon$, the norm $p$ used in adversarial training as well as the variance of the Gaussian noise.
The results are shown in Figure \ref{fig:noise}.
\vspace{-1.7 em}
We observe that the clean (i.e. in distribution; $\sigma_Q=0$) loss of the ``core" model may be higher or lower than that of the ``total" model, but in both cases, the total model is consistently more vulnerable to distributional shifts resulted from breaking spurious correlations. 

\label{sec:experiments}
\section{Empirical Evidence}

We now demonstrate increased spurious feature reliance in adversarially trained models over multiple benchmarks. We evaluate models on two backbones (ResNet18, ResNet50) adversarially trained on ImageNet \cite{imagenet} using two norms ($\ell_2, \ell_\infty$) under five attack budgets (denoted $\epsilon$) per norm, resulting in a $2\times 2\times 5 = 20$ model test suite, as well as standardly trained baselines. See appendix for details. 

\subsection{AT hurts Natural Distributional Robustness \emph{only} when Spurious Correlations are broken}

\begin{figure}
    \centering
    \includegraphics[width=0.85\linewidth]{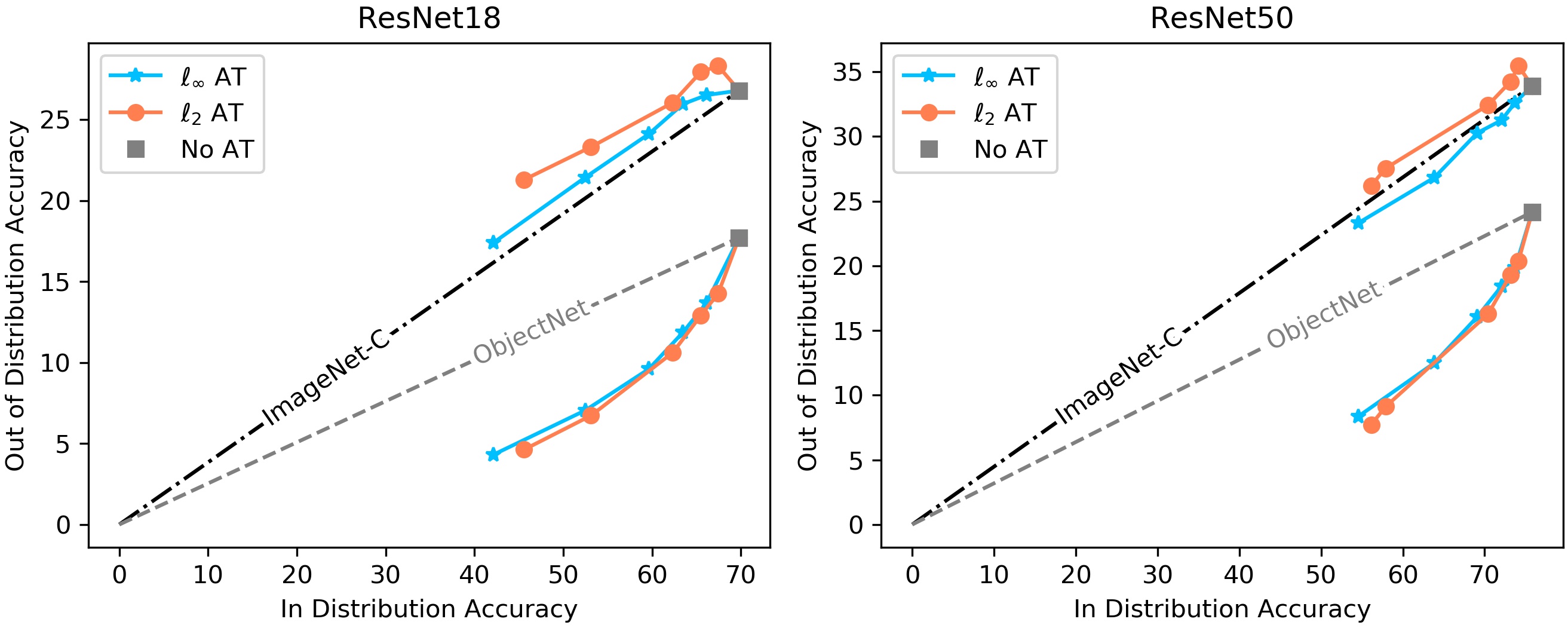}
    \caption{OOD accuracy vs standard ImageNet accuracy for adversarially trained ResNets. ImageNet-C accuracy is closely tied to standard accuracy, but for ObjectNet, where spurious correlations are broken, performance drop is more severe than a linear relation with standard accuracy would entail.}
    \label{fig:ood_vs_id}
\end{figure}

We first show reduced distributional robustness of adversarial models occurs specifically in cases where natural spurious correlations are broken. We appeal to the {\bf ImageNet-C} \cite{imagenet_c} and {\bf ObjectNet} \cite{ObjectNet} OOD benchmarks. ImageNet-C augments ImageNet samples with common corruptions like noise or blurring, distorting both core and spurious features equally. Crucially, these corruptions do not break spurious correlations. On the other hand, ObjectNet is formed by having workers capture images of common household objects (including samples from 113 classes of ImageNet) {\it in their homes}. Thus, only spurious features are affected. Namely, ObjectNet introduces distribution shifts in background, rotation, and viewpoint. We plot accuracies on these benchmarks in figure \ref{fig:ood_vs_id}.  

Recall that adversarially trained models have lower standard accuracy, which can confound our analysis, so we compare the drop in OOD accuracy to the drop in ImageNet accuracy across our model suite. Observe that the ratio of ImageNet-C accuracy to ImageNet accuracy is roughly constant across models. However, the ratio of ObjectNet accuracy to ImageNet accuracy is {\it lower} for adversarially trained models. Therefore, even after controlling for reduced standard accuracy, the distributional robustness of adversarially trained models is worse than that of standard models. Importantly, this effect does not hold for distribution shifts that maintain spurious correlations, indicating that the reduced distributional robustness is due to increased spurious feature reliance. 


\subsection{Reduced Core Sensitivity, and Difference in the Effect of $\ell_2$ \& $\ell_\infty$ Adversarial Training}
\begin{figure}[H]
    \centering
    \begin{minipage}{.47\textwidth}
    \centering
    \includegraphics[width=0.98\linewidth]{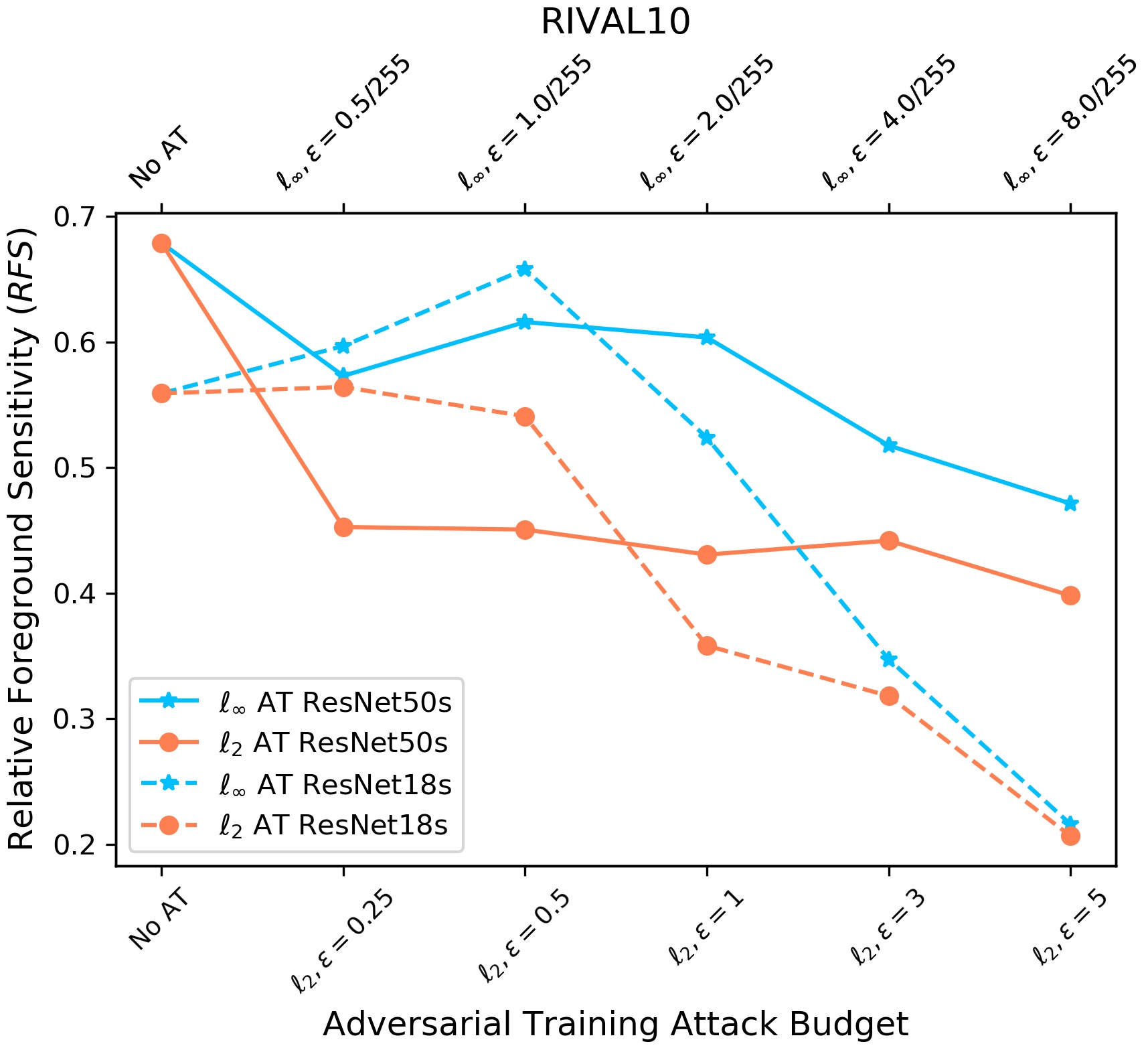}
    \end{minipage}
    \begin{minipage}{.47\textwidth}
    \centering
    \includegraphics[width=0.98\linewidth]{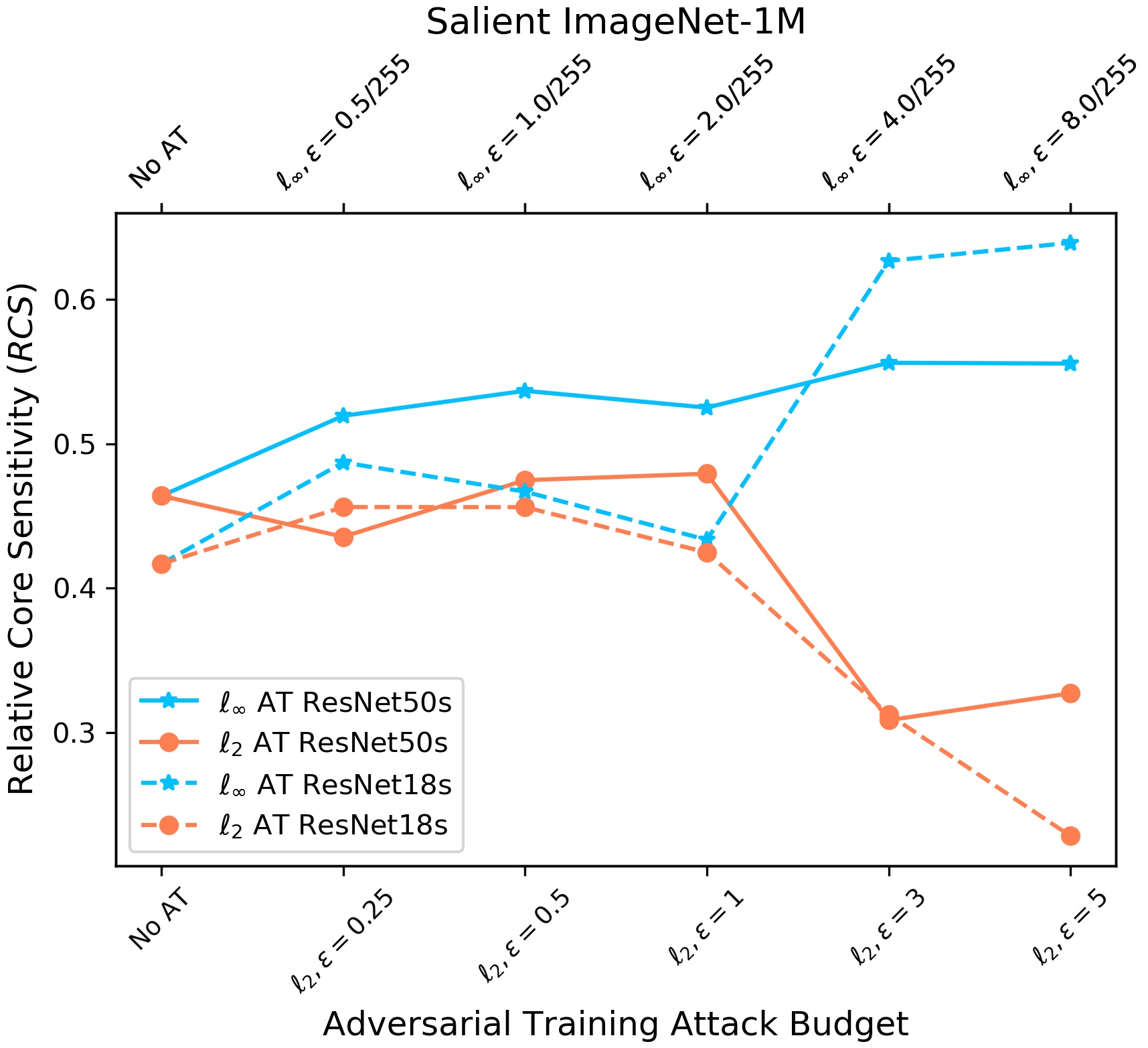}
    \end{minipage}
    \caption{Noise-based evaluation of model sensitivity to foreground ($RFS$ on RIVAL10) or core ($RCS$ on Salient ImageNet-1M) regions. Lower values entail greater sensitivity to spurious regions.}
    \label{fig:rcs}
\end{figure}

We now directly quantify sensitivity to core features via {\bf RIVAL10} and {\bf Salient ImageNet-1M} datasets \cite{rival10, salientImageNet1M}. The premise of this analysis is that model sensitivity to an input region can be quantified by the drop in accuracy due to corrupting that region \cite{iclr_salientImageNet}. \cite{rival10} introduced the noise-based metric {\it relative foreground sensitivity} ($RFS$), which is the gap between accuracy drops due to background and foreground noise, normalized so to allow for comparisons across models with varying general noise robustness. RIVAL10 object segmentations allow for $RFS$ computation. Analogously, {\it relative core sensitivity} ($RCS$) is computed using Salient ImageNet-1M's soft segmentations of core input regions. A key distinction between the two metrics is that $RCS$ is computed directly on pretrained models performing the original $1000$-way ImageNet classification task, while $RFS$ first requires models to be finetuned on the {\bf much coarser} $10$-way classification task of RIVAL10. Also, Salient ImageNet-1M includes {\it all ImageNet images}, while RIVAL10 only consists of $20$ ImageNet classes.

Figure \ref{fig:rcs} shows a decrease in $RFS$ and $RCS$ as the attack budget $\epsilon$ seen during $\ell_2$ adversarial training rises. Thus, {\bf adversarial training reduces core feature sensitivity relative to spurious feature sensitivity}. Notably, this effect does {\it not} hold for models adversarially trained with attacks under the $\ell_\infty$ norm for $RCS$, though it does for $RFS$. Alluding to our theoretical result, we conjecture that in Salient ImageNet-1M, the scales of the spurious features are much smaller than in RIVAL10, due to the diversity of images and finer grain of classes. That is, a smaller perturbation is needed to alter a spurious feature so that it correlates with an incorrect class when there are $1000$ classes than when there are only $10$ classes with generally disparate backgrounds. 



\subsection{Adversarial Training Increases Background Reliance in Synthetic Datasets}
 
Now, we take a closer look at the reliance of adversarially trained models on the contextual spurious feature of {\it backgrounds} via the synthetic datasets {\bf ImageNet-9} \cite{noise_or_signal} and {\bf Waterbirds} \cite{dro}. Both datasets use segmentations to superimpose objects over varying backgrounds, detailed below.

\begin{figure}[H]
    \centering
    \includegraphics[width=0.9\linewidth]{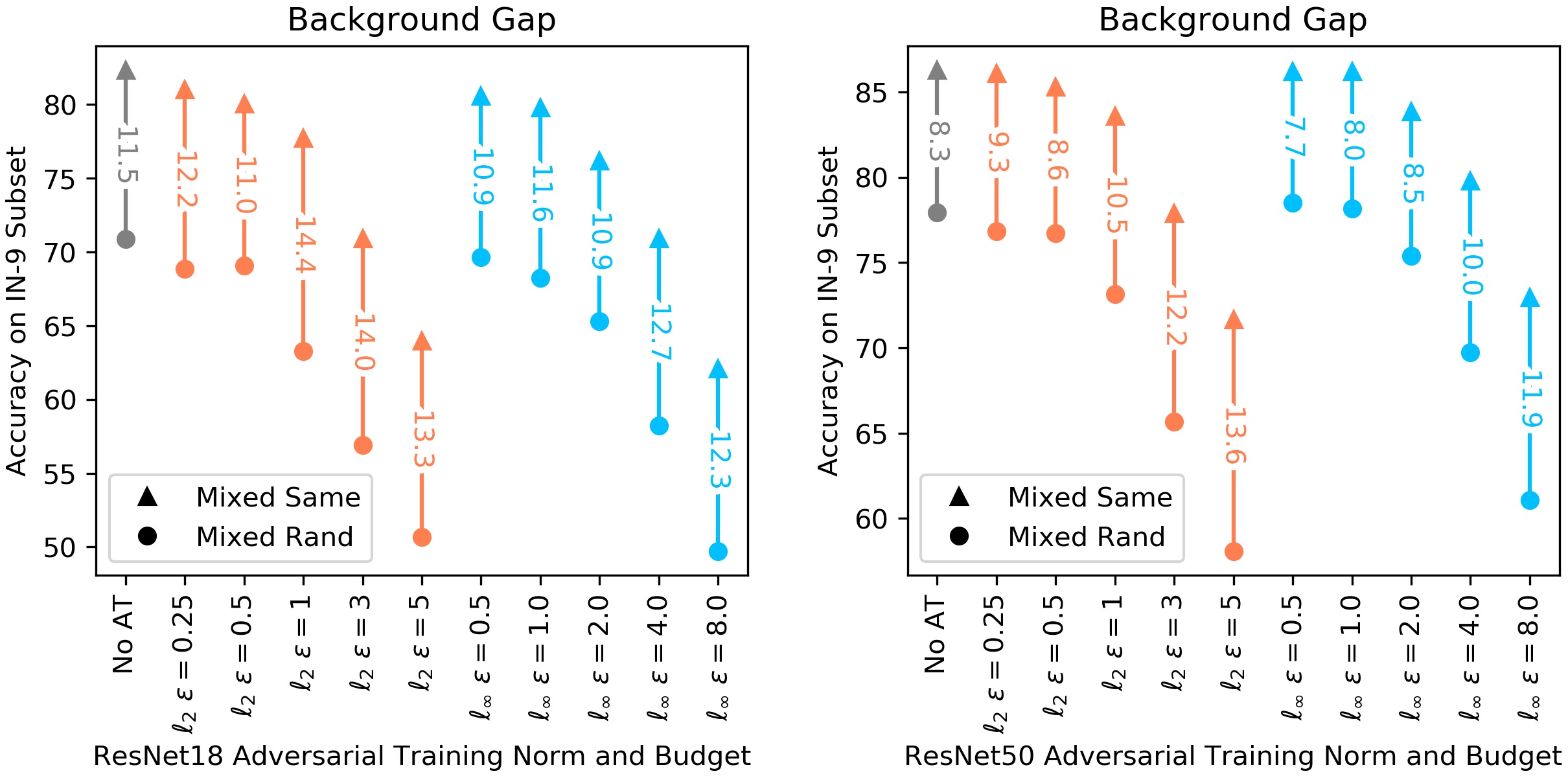}
    \caption{Background Gap (difference in accuracies on ImageNet-9 subsets $\textsc{Mixed-Same}$ and $\textsc{Mixed-Rand}$). The drop in accuracy due to background cross-class swapping ($\textsc{Mixed-Rand}$) causes larger drops in accuracy for robust models, especially $\ell_2$ adversarially trained models.}
    \label{fig:in9_bg_gap}
\end{figure}
\looseness=-1
{\bf ImageNet-9 (IN-9)} organizes a subset of ImageNet into nine  superclasses. Multiple validation sets exist for IN-9, where backgrounds or foregrounds are altered; we use $\textsc{Mixed-Same}$ and $\textsc{Mixed-Rand}$. In both sets, original backgrounds are swapped out for new ones. Crucially, in $\textsc{Mixed-Same}$, the new backgrounds are taken from other instances {\it within the same class}, while $\textsc{Mixed-Rand}$ uses {\it random backgrounds}. The metric, {\bf Background Gap}, is the difference in model accuracy on $\textsc{Mixed-Same}$ and $\textsc{Mixed-Rand}$ (i.e. drop due to breaking spurious background correlation). 

Figure \ref{fig:in9_bg_gap} shows that the background gap for our test suite of twenty robust ResNets and two standardly trained baselines. Nine out of the ten $\ell_2$ adversarially trained models have larger background gaps than the standard baselines, while the same is true for five out of the ten $\ell_\infty$ adversarially trained models. When considering relative gaps (i.e. as a percent of the accuracy on \textsc{Mixed-Same}), the increase in gap becomes even more dramatic, with the $\ell_2$ adversarially trained ResNet50 for $\epsilon=5$ having a $18.9\%$ relative drop, compared to the $9.7\%$ relative drop for the standardly trained ResNet50. 
 
\begin{figure}
    \centering
    \includegraphics[width=0.9\linewidth]{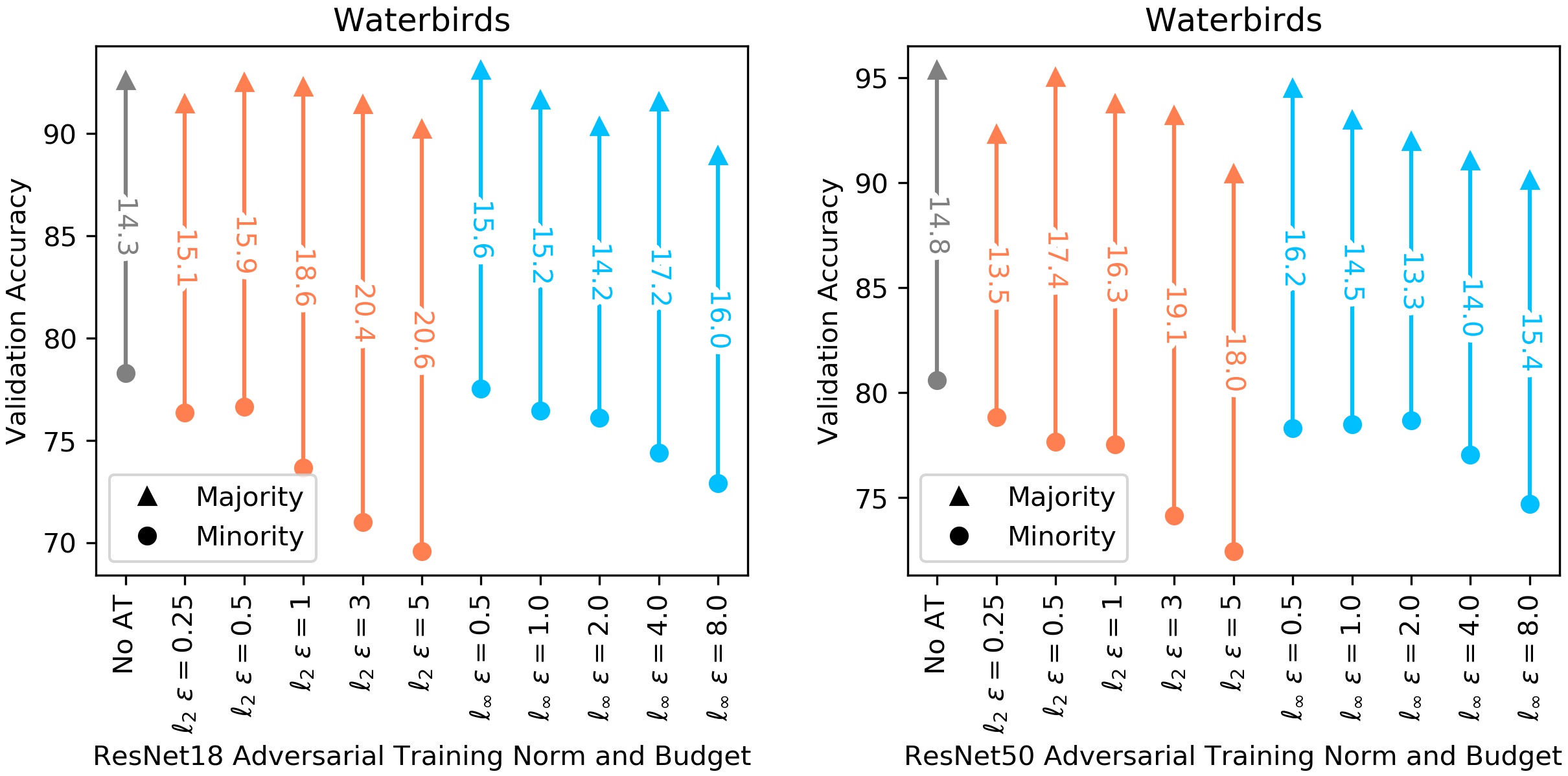}
    \caption{Accuracies on Waterbirds subsets where spurious correlation is intact (majority group; e.g. land birds on land backgrounds) and where it is broken (minority group; e.g. land birds on water backgrounds). The drop in performance due to breaking the background correlation grows larger for robust models, especially $\ell_2$ adversarially trained models.}
    \label{fig:waterbirds}
\end{figure} 
\textbf{Waterbirds} combines foregrounds from Caltech Birds  \cite{cub} and backgrounds from SUN Places \cite{sun}. The task is binary classification of landbirds and waterbirds. The majority group ($95\%$ of training samples) consists of landbirds over land backgrounds and waterbirds over water backgrounds. The minority group breaks this spurious correlation, placing landbirds over water backgrounds, and vice versa. The test set is evenly split between these groups. We train only a final linear layer atop the frozen feature extractors (so that models remain adversarially robust) for each of our models on the Waterbirds training set for ten epochs, saving the model with highest validation accuracy. 

Figure \ref{fig:waterbirds} shows majority and minority group accuracies, and the gap between them, for our test suite of models. Again, we see increased gaps for robust models, with $100\%$ of $\ell_2$ and $60\%$ of $\ell_\infty$ models respectively having larger gaps than the standardly trained baseline on the corresponding backbone. 

In both benchmarks, breaking the background spurious correlation causes a more significant drop in performance for adversarially trained models than standardly trained models, indicating that adversarial training led to increased reliance on backgrounds. The observed affect is stronger for $\ell_2$ adversarially trained models than $\ell_\infty$ ones. Further, the gaps grow near monotonically with $\epsilon$.

\label{sec:reverse}
\subsection{Reverse Effect: Presence of Spurious Correlations Can \emph{Improve} Adversarial Robustness}
\begin{wrapfigure}{r}{4.9cm}
\centering
\includegraphics[width=0.8\linewidth]{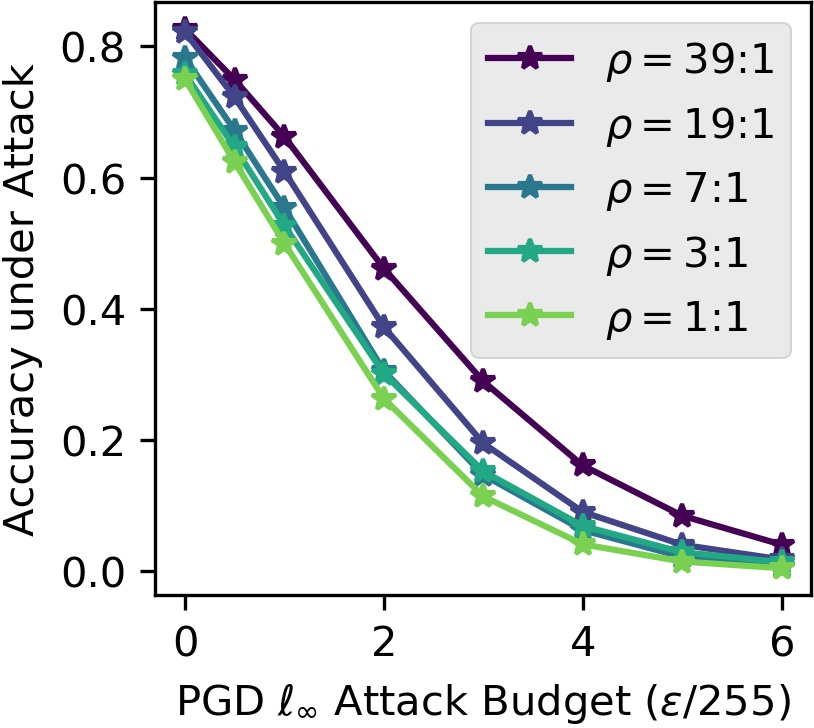}
\caption{Injecting spurious feature improves adversarial robustness.}
\label{fig:reverse}
\end{wrapfigure}
Finally, we show evidence that is directly at odds with the claim that spurious features lead to adversarial vulnerability. We train ResNet18s on CIFAR10 \cite{cifar10} with a spurious feature injected. Namely, images have all values in one color channel slightly increased. The majority group consists of red-shifted images from classes $0-4$ and green shifted images from classes $5-9$, while the minority group has reverse color-shifts. The parameter $\rho$ is the ratio between majority and minority group size, controlling the strength of the spurious correlation (higher $\rho$ means stronger spurious correlation; $\rho=1:1$ means the spurious feature has no predictive power). We then evaluate the accuracy of the trained model under adversarial attack on an i.i.d. test set ({\bf spurious feature retained}, no distribution shift). 

\looseness=-1
Figure \ref{fig:reverse} visualizes the results. Not surprisingly, the clean accuracy is higher for models trained on data with higher $\rho$, as the spurious feature is more predictive for higher $\rho$. In fact, the added predictive influence the spurious feature leads to better accuracy {\it under attack}, with the gap between highest and lowest $\rho$ values growing up to four fold compared to the baseline gap in clean accuracy. Thus, using a spurious feature can improve adversarial robustness. Despite being on a contrived example, this experiment shows that, while some spurious correlations may cause adversarial vulnerability, {\it others do the opposite}: the picture is more nuanced than previously assumed.

\section{Acknowledgements}

This project was supported in part by NSF CAREER AWARD 1942230, a grant from NIST 60NANB20D134, HR001119S0026 (GARD), ONR YIP award N00014-22-1-2271, Army Grant No. W911NF2120076 and the NSF award CCF2212458.

\bibliographystyle{abbrv}
\bibliography{main}

\input{appendix}

\end{document}

%% file: appendix.tex
\clearpage

\appendix

\section{Limitations of Theorem \ref{thm:only}}
While Theorem \ref{thm:only} provides an understanding of the tradeoff between adversarial and natural distributional robustness, there are some limitations.
Firstly, the results consider a setting where the core and spurious features are completely disentangled, i.e, they each represent different parts of the input. In practice, spurious features may be entangled with the core features (e.g., the color of an image may represent a spurious feature.) 
In addition, our results mainly consider the \emph{goal} of adversarial training as we focus on the expected loss $L_{p, \epsilon}(\theta)$, rather than its finite-sample variant.
This is because even for an $\ell_2$ adversary,
characterizing the finite-sample behaviour of adversarial training is difficult and requires careful assumptions on the asymptotic behaviour of the parameters (e.g., see Theorem 3.3 in \cite{javanmard2020precise}). 
We leave exploring these directions to future work. Even so, we believe our theoretical results are of interest to the community since disjoint features already capture
a wide variety of spurious correlations, e.g., background correlations, as well as examples where a spurious object is present in the image. The main goal of our theoretical analysis is to show the existence of explicit tradeoffs between adversarial and distributional robustness and build practical insights using those results.

\section{Societal Impact}

Our work touches on two important notions of robustness for the safe and fair deployment of deep models in the wild. We hope our results lead to careful analysis of all modes of robustness, and the interplay between them, before deep models are used in sensitive applications. While our results create tension with some previous works \cite{cama, causaladv, ftrs_not_bugs}, we stress that we do not wish to diminish their work; instead, we hope our work reveals the vast nuance associated with spurious correlations, which can help and hurt models in various ways. Lastly, we release all code to encourage future work.
 
\section{Varying the Number of Core and Spurious Features}

In this section, we further analyze the plateauing behaviour of the performance of the linear model observed in Figure \ref{fig:norm_frac} 
for different values of core and spurious features with the $\ell_1$ and $\ell_2$ adversarial training.

\begin{figure}[ht]
    \centering
	\begin{subfigure}{0.4\textwidth}
    	\centering
		\resizebox{0.98\linewidth}{!}{\includegraphics{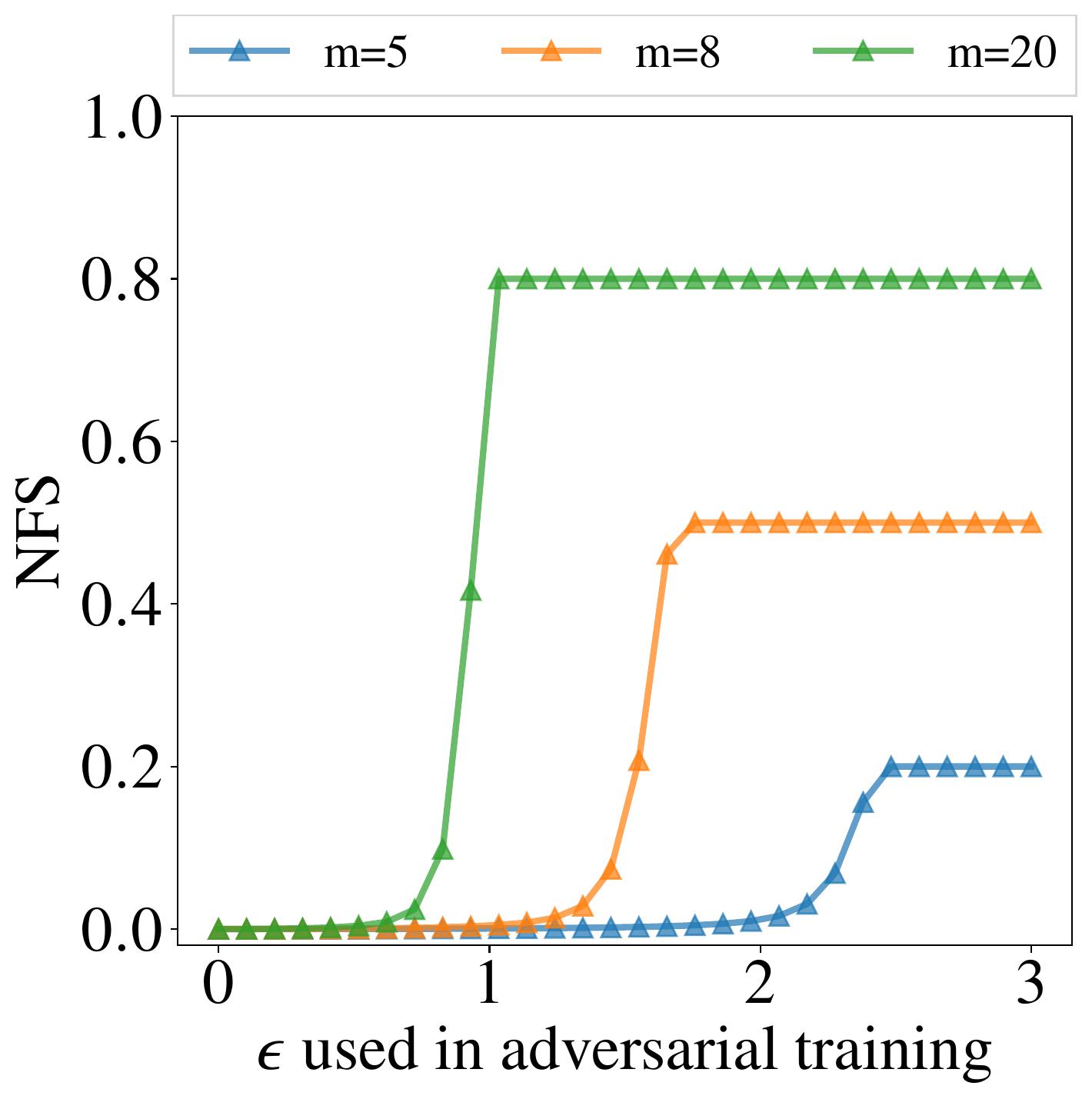}}
		\caption{$4$ core feautres}
	\end{subfigure}
    \begin{subfigure}{0.4\textwidth}
    	\centering
		\resizebox{0.98\linewidth}{!}{\includegraphics{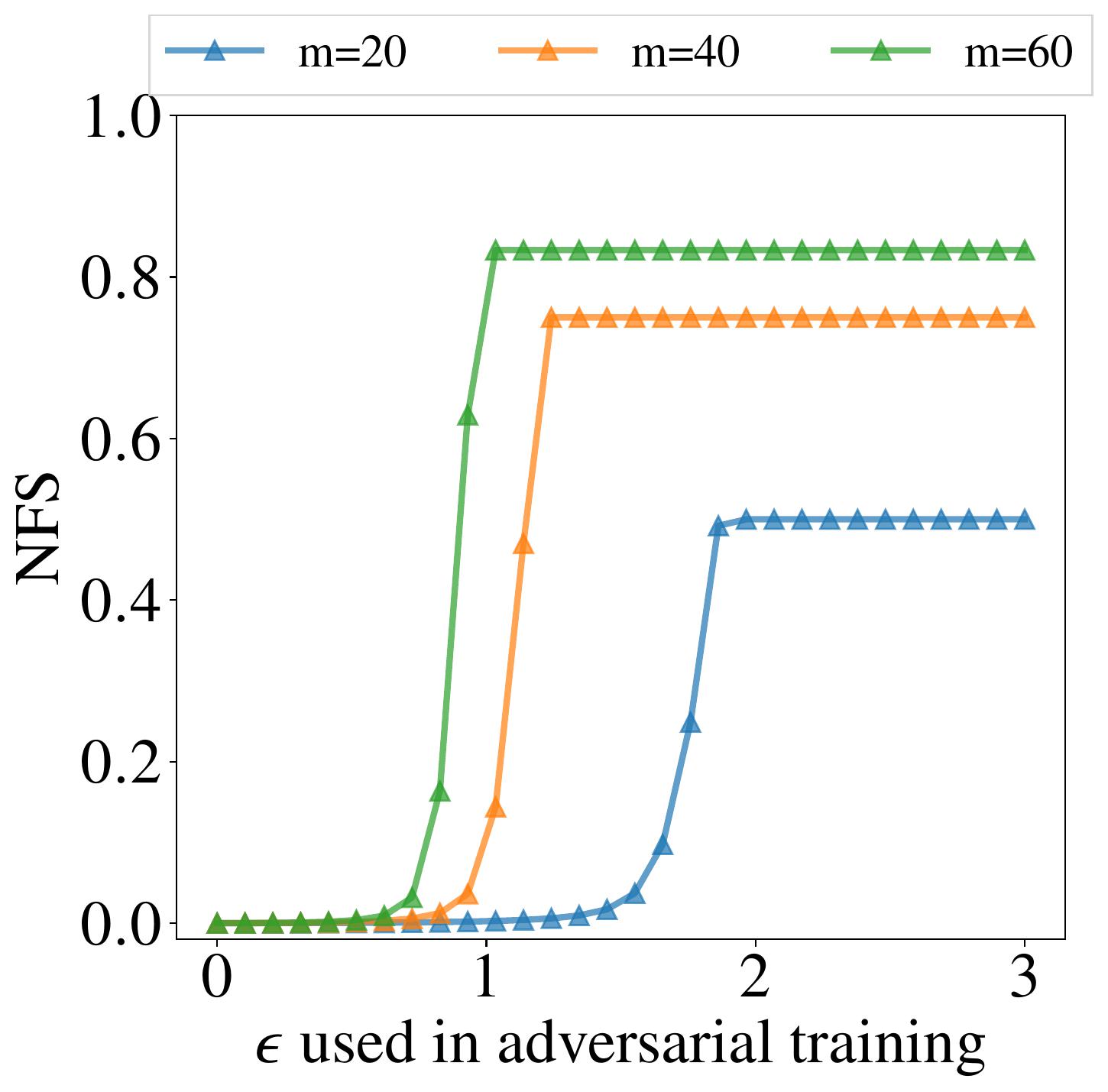}}
		\caption{$10$ core feautres}
	\end{subfigure}
    \caption{
    Analysis of \nfs{} for the linear model when using the $\ell_1$ norm in adversarial training. Each figure measures the reliance of the model on spurious features (measured by \nfs{})
    while varying the adversarial training budget $\epsilon$, using different number of \emph{total} features $m$. 
    The number of core features is kept constant and set to $4$ in Figure \textbf{(a)} and to $10$ in
    Figure \textbf{(b)}.
    }
    \label{fig:norm_frac_2}
\end{figure}

We first focus on
the $\ell_1$ case.
We consider different values for the number of core features $c$ and total features $m$
and measure \nfs{} for different values of adversarial budget $\epsilon$ as in Figure \ref{fig:norm_frac}. 
The matrix
$\Sigma$ is constructed using Equation \eqref{eq:def_q} as before, with modified number of rows and columns based on the values of $c, p$.
Similarly, $\opttheta$ is constructed as before, with the core coordinates set to $1$ and the spurious coordinates set to $0$.
The value of $\eta$ is fixed at $0.5$.
The results are shown in Figure \ref{fig:norm_frac_2}. 

As shown in the Figure, when using $m$ total features and $c$ core features, \nfs{} plateaus at
$\frac{m-c}{m}$ for large values of $\epsilon$. 
Intuitively, this is because of the structure of the optimization problem \eqref{eq:thm_statement}.
Recall that when using the $\ell_1$ norm, the value
of $q$ in \eqref{eq:thm_statement} equals $\infty$. As such, adversarial training tries to find  a parameter $\theta$ that has a low $\ell_{\infty}$ norm and is ``close'' (as measured by $\sigma_{\theta}$) to $\opttheta$. The $\ell_\infty$ penalty encourages values of $\theta$ that are uniform across the coordinates. 
Since there are $m-c$ spurious features and $m$ total features, this leads to models that have an \nfs{} value of $\frac{m-c}{m}$.

\begin{figure}[ht]
    \centering
	\begin{subfigure}{0.4\textwidth}
    	\centering
		\resizebox{0.98\linewidth}{!}{\includegraphics{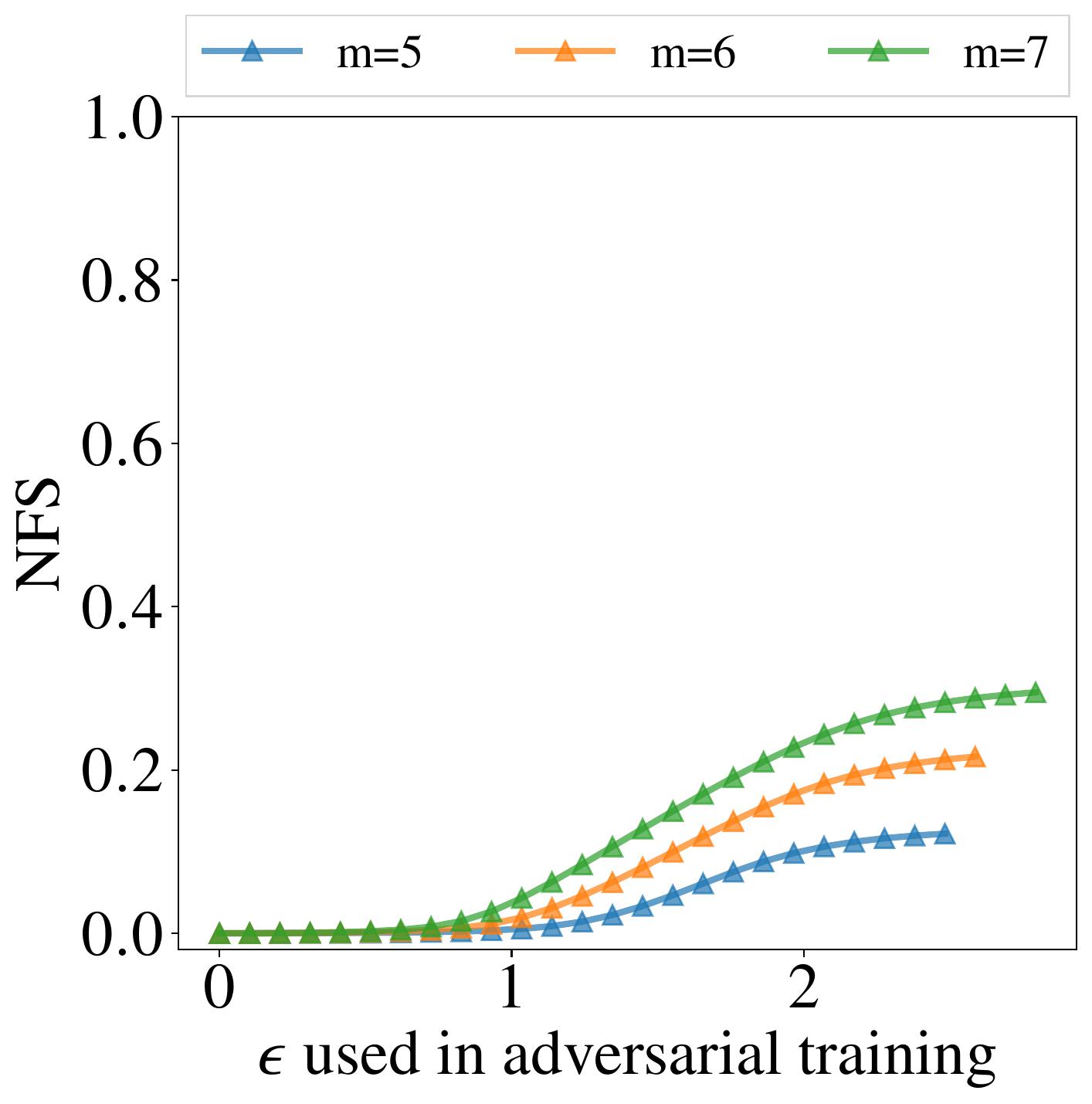}}
		\caption{$4$ core feautres}
	\end{subfigure}
    \begin{subfigure}{0.4\textwidth}
    	\centering
		\resizebox{0.98\linewidth}{!}{\includegraphics{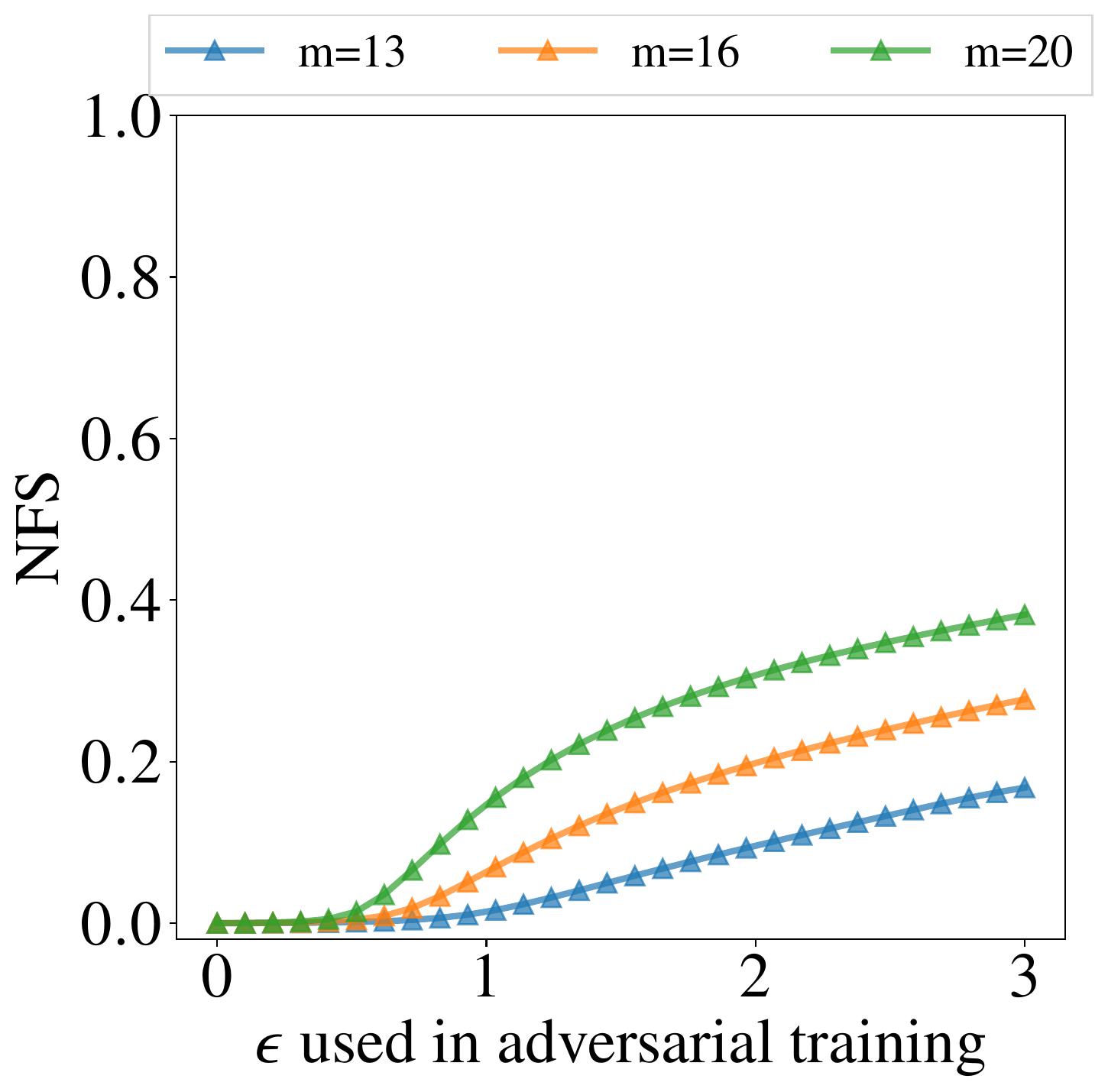}}
		\caption{$10$ core feautres}
	\end{subfigure}
    \caption{
    Analysis of \nfs{} for the linear model when using the $\ell_2$ norm in adversarial training. Each figure measures the reliance of the model on spurious features (measured by \nfs{})
    while varying the adversarial training budget $\epsilon$, using different number of \emph{total} features $m$. 
    The number of core features is kept constant and set to $4$ in Figure \textbf{(a)} and to $10$ in
    Figure \textbf{(b)}.
    }
    \label{fig:norm_frac_3}
\end{figure}

We further repeat the above experiment with $\ell_2$ norm. The results are shown in Figure \ref{fig:norm_frac_3}.
As seen in the figure, we see the same qualitative results as in Figure \ref{fig:norm_frac}, with higher values of \nfs{} when increasing numbers of spurious features. 

\section{Proof of Theorem \ref{thm:only}}

\begin{proof}
  We first claim that
  \begin{align*}
    \max_{\norm{\delta} \le \epsilon} (Y - \vecdot{X + \delta}{\theta})^2
     = \left(
       \abs{Y - \vecdot{X}{\theta}} + \epsilon \cdot \norm{\theta}_q
     \right)^2
     \label{eq:loss_identity_dual_norm_2}
  \end{align*}
  To see why this holds, note that for all $\delta$ satisfying
  $\norm{\delta}_p \le \epsilon$,
  \begin{align*}
    \abs{Y - \vecdot{X + \delta}{\theta}}
    &\overset{(a)}{\le}
    \abs{Y - \vecdot{X}{\theta}} + \abs{\vecdot{\delta}{\theta}}
    \\& \overset{(b)}{\le}
    \abs{Y - \vecdot{X}{\theta}} + \epsilon \cdot \norm{\theta}_q,
  \end{align*}
  where $(a)$ follows from the triangle inequality and $(b)$ follows from
  Hölder's inequality. With a suitable choice of $\delta$,
  we can achieve equality for $(b)$. As $\norm{\theta}_q = \norm{-\theta}_{q}$,
  at least one of $\{\delta, -\delta\}$ would further achieve equality for $(a)$. 
  As maximizing $\abs{.}$ is equivalent to maximizing $\left( . \right)^2$, \eqref{eq:loss_identity_dual_norm} is proved.

  Given \eqref{eq:loss_identity_dual_norm}, we can rewrite \eqref{eq:loss_adv}
  as
  \begin{align*}
    L_{p, \epsilon} &= \Ex{
      \parant{
        \abs{ Y - \vecdot{X}{\theta}} + \epsilon \cdot \norm{\theta}_q
      }^2
    }
    \\&=
    \Ex{
      \parant{
      Y - \vecdot{X}{\theta}
      }^2
    }
    + \epsilon^2 \cdot \norm{\theta}_q^2  + 2\cdot \epsilon \cdot \norm{\theta}_q\cdot \Ex{\abs{Y - \vecdot{X}{\theta}}}
    \\&\overset{(a)}{=}
    \Ex{
      \parant{
        \vecdot{X}{\theta - \opttheta}
        + W
      }^2
    }
    + \epsilon^2 \cdot \norm{\theta}_q^2  + 2\cdot \epsilon \cdot \norm{\theta}_q \cdot \Ex{\abs{\vecdot{X}{\theta - \opttheta} + W}},
  \end{align*}
  Where for $(a)$ we have used the fact that $Y = \vecdot{X}{\theta} + W$.
  
  Define $v_{\theta}$ as $\vecdot{X}{\theta - \opttheta} + W$.
  As $X$ was assumed to be sampled from $\mN(0, \Sigma)$, $v_{\theta}$ is distributed as $\mN(0, \sigma_{\theta}^2)$.
  It follows that
  \begin{align*}
    L_{p, \epsilon} &= 
    \Ex{
      \parant{
        \vecdot{X}{\theta - \opttheta}
        + W
      }^2
    }
    + \epsilon^2 \cdot \norm{\theta}_q^2  + 2\cdot \epsilon \cdot \norm{\theta}_q \cdot \Ex{\abs{\vecdot{X}{\theta - \opttheta} + W}},
    \\&=
    \Ex{v_{\theta}^2} + \epsilon^2 \cdot \norm{\theta}_q^2 + 2 \cdot \epsilon\cdot \norm{\theta}_q \cdot \Ex{\abs{v_{\theta}}}
    \\&\overset{(a)}{=}
    \sigma_{\theta}^2 + \epsilon^2 \cdot \norm{\theta}_q^2 + 2 \cdot c_1 \cdot \epsilon \cdot \norm{\theta}_q \cdot \sigma_{\theta}
    \\&=
    (c_1^2 + c_2) \cdot \sigma_{\theta}^2 + 
    \epsilon^2 \cdot \norm{\theta}_q^2 + 2 \cdot c_1 \cdot\epsilon \cdot \norm{\theta}_q \cdot \sigma_{\theta}
    \\&=
    c_2 \cdot \sigma_{\theta}^2 + (c_1\sigma_{\theta} + \epsilon \cdot \norm{\theta}_q)^2
  \end{align*}
  where for $(a)$ we have used the fact that $\Ex{\abs{\mN(0, \sigma^2}} = c_1 \cdot \sigma$.
  This proves \eqref{eq:thm_statement} as claimed.
  ~\\
  As for convexity, $\sigma_{\theta}$ is convex in $\theta$ since it can be written as
  $\norm{[\Sigma^{\frac{1}{2}}(\theta - \opttheta), \sigma_{w}]}_2$ where
  $[., .]$ denotes the vector stacking operation.
  As $c_1\sigma_{\theta} + \epsilon \cdot \norm{\theta}_q$ is always positive
  and $x \to x^2$ is convex and increasing for $x \ge 0$,
  this implies that 
  $(c_1\sigma_{\theta} + \epsilon \cdot \norm{\theta}_q)^2$ is convex as well.
  Finally $c_2 \sigma_\theta^2$ is convex as $c_2 > 0$ and therefore \eqref{eq:loss_adv} is convex in $\theta$.
\end{proof}

\section{Additional Details on Reverse Effect (Section \ref{sec:reverse})}

Our final empirical observation is that the presence of a spurious feature (in both training and test distributions) can lead to increased adversarial robustness. This more directly creates tension with claims that adversarial vulnerability is born out of spurious feature reliance. We refer to this as the `reverse effect', in relation to our primary empirical and theoretical finding that adversarial training increases spurious feature reliance. We now elaborate on the experimental setup discussed in Section \ref{sec:reverse}, reproduce the results with a different spurious feature, and finally appeal to ImageNet-9 to demonstrate this effect using a more realistic spurious feature (i.e. backgrounds).

\subsection{Experimental Setup}

\begin{figure}
    \centering
    \begin{minipage}{0.42\linewidth}
    \centering
    \includegraphics[width=0.47\textwidth]{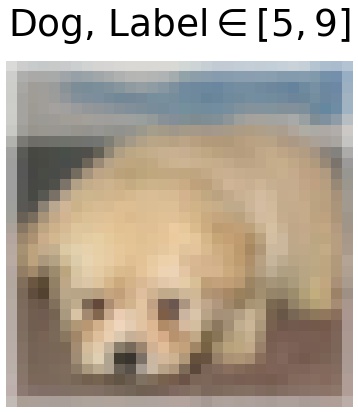} \\
    \includegraphics[width=0.98\textwidth]{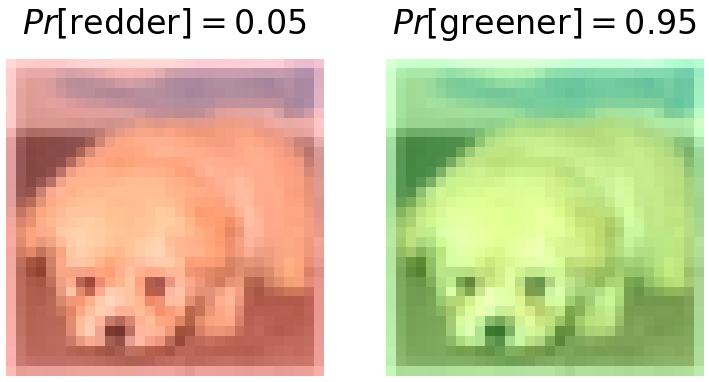}
    \caption{Color Shift, $\rho=19:1$}
    \label{fig:spur_corr1}
    \end{minipage}
    \begin{minipage}{0.42\linewidth}
    \centering
    \includegraphics[width=0.47\textwidth]{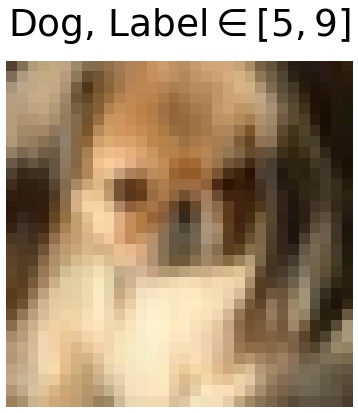} \\
    \includegraphics[width=0.98\textwidth]{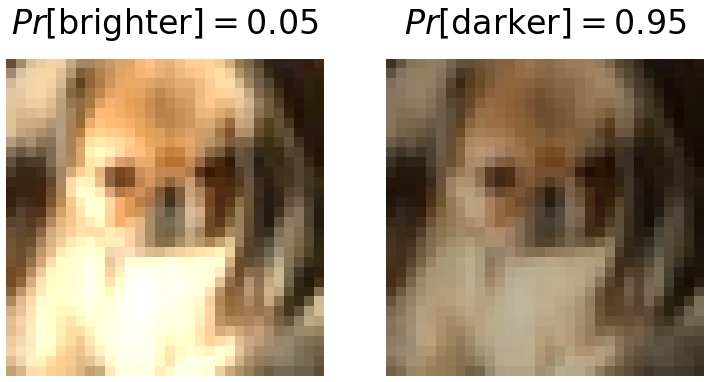}
    \caption{Lighting Shift, $\rho=19:1$}
    \label{fig:spur_corr2}
    \end{minipage}
\end{figure}
\noindent{{\bf Overview.}} We inject spurious correlations to the CIFAR10 dataset. Based on the class label, we adjust half the images (i.e. with class label from 5 to 9) to shift in one direction with high probability. For example, a dog image is made greener with probability $0.95$, corresponding to a majority-to-minority group ratio of $\rho=19:1$. With probability $0.05$, we shift in the other direction (e.g. make redder). We then standardly train a ResNet18 from scratch on the dataset with the spurious feature injected for the 10-way CIFAR classification task. Importantly, we evaluate robust accuracy with the spurious feature retained, and then compare adversarial robustness of models trained under data with different strengths of the injected spurious correlation.
\begin{wrapfigure}{r}{5.2cm}
\centering
\includegraphics[width=0.8\linewidth]{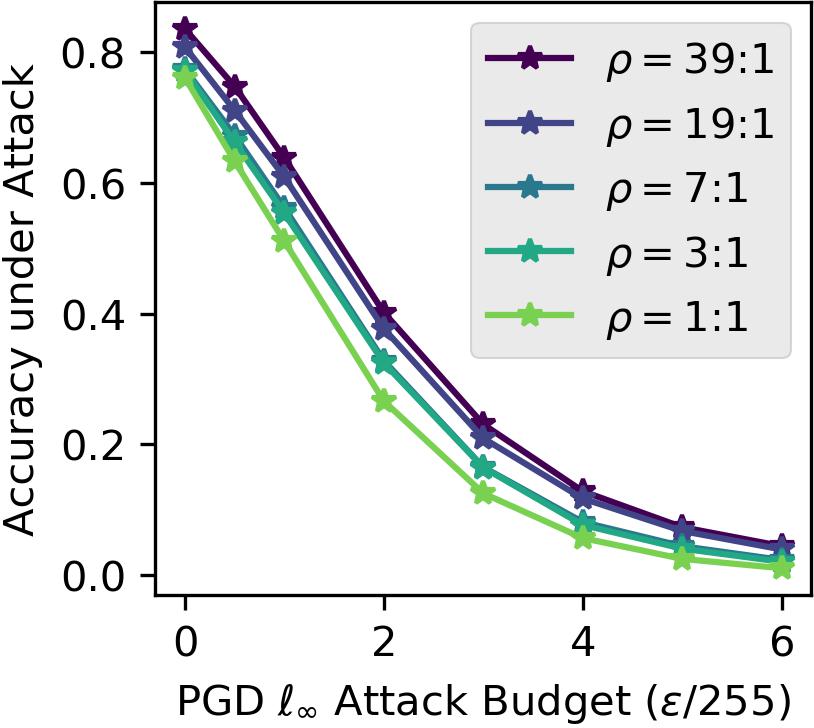}
\caption{Reverse effect using spurious feature of {\it lighting}. Main text figure uses color as spurious feature.}
\label{fig:reverse2}
\end{wrapfigure}
Figure \ref{fig:reverse} and \ref{fig:reverse2} show that for two distinct spurious features (color and lighting), robust accuracy is higher when the spurious correlation is stronger. Notably, the gain is larger than the gain in standard accuracy. Intuitively, we see that relying on the predictive power of the spurious feature is helpful for standard accuracy, and especially for acccuracy under adversarial attack. Despite being irrelevant to the true labeling function, the spurious feature can improve model performance, and indeed even lead to better adversarial robustness. 

\noindent{{\bf Details.}} Color shift is achieved by increasing all pixel intensities along one channel by $0.25$. Lighting shift is achieved by simply scaling an input by $1.25$ to make brighter or $0.75$ to make darker. All images are clamped to remain in the $[0,1]$ pixel range after spurious feature injection. Models are trained for 20 epochs using an Adam optimizer with a learning rate of $0.001$ and weight decay of $1e-4$. 

\subsection{Leveraging ImageNet-9}

We now demonstrate the observed reverse effect on the higher resolution ImageNet-9 dataset, leveraging the natural and ubiquitous spurious feature of backgrounds. We finetune pretrained models on \textsc{Mixed-Same} and \textsc{Mixed-Rand} separately, and evaluate each model's accuracy under attack on the same split that they were trained over. Further, we leverage the adversarially trained models from test suite in this experiment. This way, accuracy under attack is more informative, as the models are trained to expect attacks (i.e. we are not imposing any distribution shifts that would lead to unexpected model behavior). Along this vain, we attack each backbone with the same norm and $\epsilon$ that it was pretrained over.

\begin{figure}
    \centering
    \includegraphics[width=0.95\linewidth]{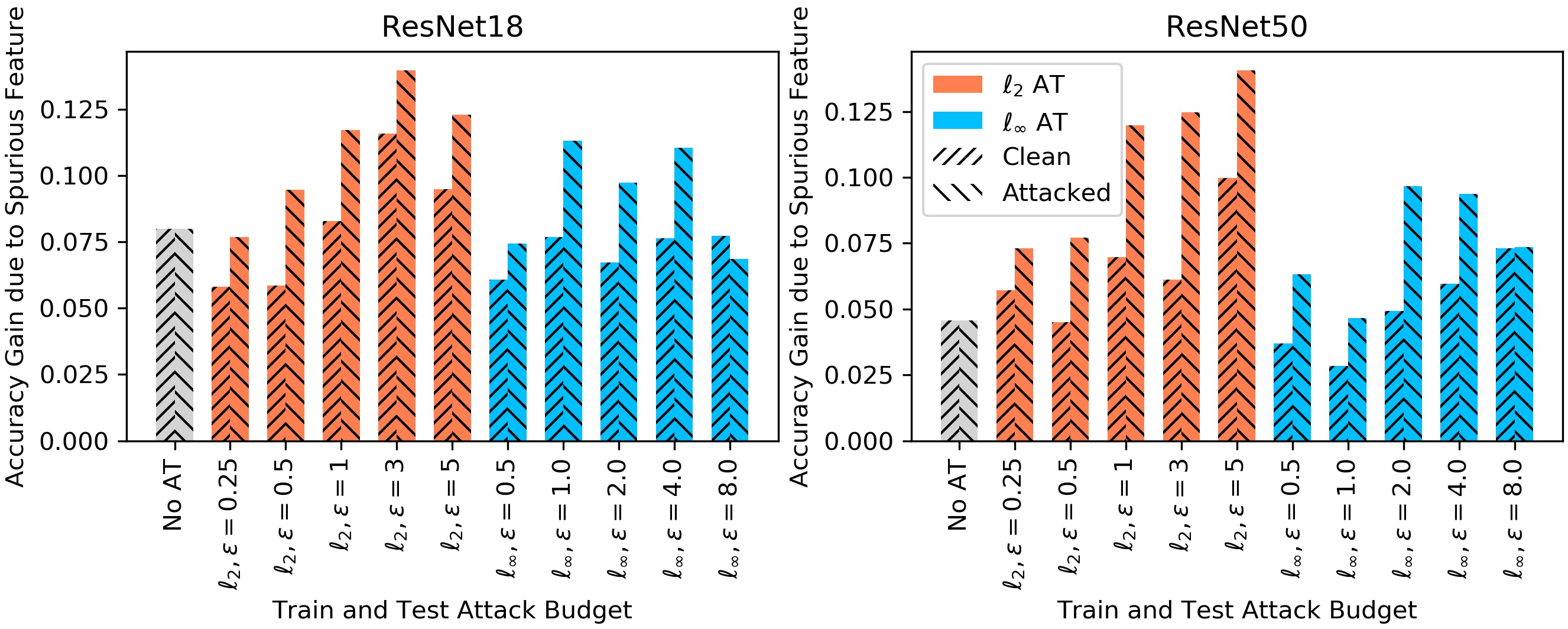}
    \caption{Background Gap (difference in accuracy on \textsc{Mixed-Same} and \textsc{Mixed-Rand}) for clean and adversarially attacked images. Across models, background gap is larger when considering accuracy under attack, suggesting that the presence of a spurious correlation in training data makes the model more adversarially robust over the same distribution.}
    \label{fig:ms_mr_diff}
\end{figure}
Figure \ref{fig:ms_mr_diff} shows the gain in accuracy for the models trained and evaluated on \textsc{Mixed-Same} compared to those using \textsc{Mixed-Rand}. We see that the presence of background correlations increases both standard and robust accuracy for all models (i.e. gains are positive). Further, gains in accuracy under attack are larger than gains in standard accuracy in nearly all cases. Thus, it seems like the added predictive power of the spurious background feature has a significantly nontrivial impact on improving adversarial robustness, contradicting many existing arguments on the link between spurious correlations and adversarial vulnerability.

\section{Adversarially Robust Model Test Suite (Section \ref{sec:experiments})}

\subsection{Model Details}

We utilize the treasure trove of open-source adversarially trained models, contributed by \cite{robust_models_transfer}, accessible at \url{https://github.com/Microsoft/robust-models-transfer}. For completeness, we now provide details on the models we use, though we refer readers to Appendix A.1 of the original text, where the information we share now is sourced. 

\noindent{\textbf{Training}} All models were trained on ImageNet in batches of $512$ samples, using SGD optimizer with momentum of $0.9$ and weight decay of $1e-4$, for a total of $90$ epochs, with learning rate dropping by a factor of $10$ every $30$ epochs. The standard procedure of \cite{pgd} was performed to adversarially train models, using $3$ projected gradient descent steps with a step size $\frac{2}{3}\epsilon$ for the attack budget $\epsilon$.

\noindent{\textbf{Selected Models}} We focus our empirical study on the ResNet architecture \cite{resnet} because of its wide spread popularity. Specifically, we study ResNet18s and ResNet50s that are adversarially trained under the $\ell_2$ norm, for $\epsilon \in \{0.25, 0.5, 1, 3, 5\}$, and $\ell_\infty$ norm, for $\epsilon \in \{0.5/255, 1/255, 2/255, 4/255, 8/255\}$, as well as standardly trained baselines. 
\begin{table}[]
    \centering
    \begin{tabular}{c|c|cc|c} \toprule
        AT Norm & $\epsilon$ & ResNet18 & ResNet50 & Wide ResNet50 (2x) \\ \midrule
        \multicolumn{2}{c|}{No Adv Training} & 69.79 & 75.80 & 76.97\\ \midrule
        $\ell_2$ &$0.25$ & 67.43 & 74.14  & 76.21\\
        $\ell_\infty$ &$0.5/255$ &  66.13 & 73.73 & 75.82\\ \hline
        $\ell_2$ &$0.5$ &  65.49 & 73.16   & 75.11\\
        $\ell_\infty$ &$1/255$ &  63.46 & 72.05& 74.65\\ \hline
        $\ell_2$ &$1$ &  62.32 & 70.43 & 73.41\\
        $\ell_\infty$ &$2/255$ &  59.63 & 69.10  &72.35 \\ \hline
        $\ell_2$ &$3$ &  53.12 & 62.83 & 66.90\\
        $\ell_\infty$ &$4/255$ &  52.49 & 63.86  & 68.41\\ \hline
        $\ell_2$ &$5$ &  45.59 & 56.13 & 60.94\\
        $\ell_\infty$ &$8/255$ &  42.11 & 54.53 & 60.82\\ \bottomrule 
    \end{tabular} \vspace{0.1cm}
    \caption{Clean ImageNet accuracy for test suite of $\ell_2$ and $\ell_\infty$ adversarially trained ResNets over varying $\epsilon$. Observe that the $i^{th}$ $\ell_2$ AT model has similar clean accuracy to the $i^{th}$ $\ell_\infty$ AT model.}
    \label{tab:standard_accs}
\end{table}
\begin{table}[]
    \centering
    \begin{tabular}{c|ccccc} \toprule
         &ShuffleNet &  MobileNet  & VGG & DenseNet & ResNeXt \\ \midrule
         No AT &  64.25 &65.26 & 73.66 & 77.37 & 77.38\\
         $\ell_2$ AT, $\epsilon=3$  & 43.32& 50.40 & 57.19 & 66.98 & 66.25\\ \bottomrule
    \end{tabular} \vspace{0.1cm}
    \caption{Clean ImageNet accuracy for five additional architectures considered. }
    \label{tab:extra_models}
\end{table}

\looseness=-1
Table \ref{tab:standard_accs} shows the standard accuracies for these models. Note that we at times compare between the $\ell_2$ and $\ell_\infty$ adversarially trained models (e.g. figure \ref{fig:rcs}). We acknowledge that direct comparisons are challenging because the threat model under which adversarial robustness is optimized for are different. However, we note that standard accuracies of the $i^{th}$ $\ell_2$ AT model is roughly the same as that of the $i^{th}$ $\ell_\infty$ AT model, suggesting that those models lie in similar points of the accuracy-robustness tradeoff. 

\noindent{{\bf Additional Models. }} We extend our analysis to other architectures. We replicate all pretrained-model experiments on the Wide ResNet50 (2x) backbones, for which we have checkpoints for each of the five $\epsilon$ values for both $\ell_2$ and $\ell_\infty$ norms. We also inspect MobileNetv2 \cite{mobilenet}, DenseNet161 \cite{densenet}, ResNeXt5050\_32x4d \cite{resnext}, ShuffleNet \cite{shufflenet}, and VGG16\_bn \cite{vgg}. For each of these five architectures, we compare an $\ell_2$ adversarially trained model with $\epsilon=3$ to a standardly trained baseline. 

\subsection{Experimental Details}

\noindent{{\bf ObjectNet and ImageNet-C}} \cite{ObjectNet, imagenet_c}. We report raw accuracies under noise, blur, and digital corruption types for ImageNet-C, as opposed to relative corruption error. For ObjectNet, we map ImageNet predictions to the set of 113 overlapping classes in ObjectNet. {\bf RIVAL10 ($RFS$) and Salient ImageNet-1M ($RCS$)} \cite{rival10, salientImageNet1M}. $RFS$ computation requires finetuning a final linear layer over fixed features for the coarse-grained ten way RIVAL10 classification. $RCS$ operates on models off the shelf, directly inspecting accuracies over ImageNet classes (and samples, with region-based noise corruption). {\bf ImageNet-9 and Waterbirds} \cite{noise_or_signal, dro}. ImageNet-9 accuracies are obtained by mapping off-the-shelf model predictions to the nine coarse labels deterministically. Waterbirds requires finetuning, which we do over fixed features. For RIVAL10 and Waterbirds finetuning, we use Adam with learning rate of $1e-4$ and weight decay of $1e-5$ for $20$ and $15$ epochs respectively. 

\subsection{Results on Extended Model Test Suite}
We now corroborate all our empirical findings on new backbones, expanding our analysis to 21 new models (including 10 AT WideResNet50s over both $\ell_2$ and $\ell_\infty$ norms) over six architectures.

\begin{wrapfigure}{r}{5.8cm}
\vspace{-0.3cm}
\centering
\includegraphics[width=0.8\linewidth]{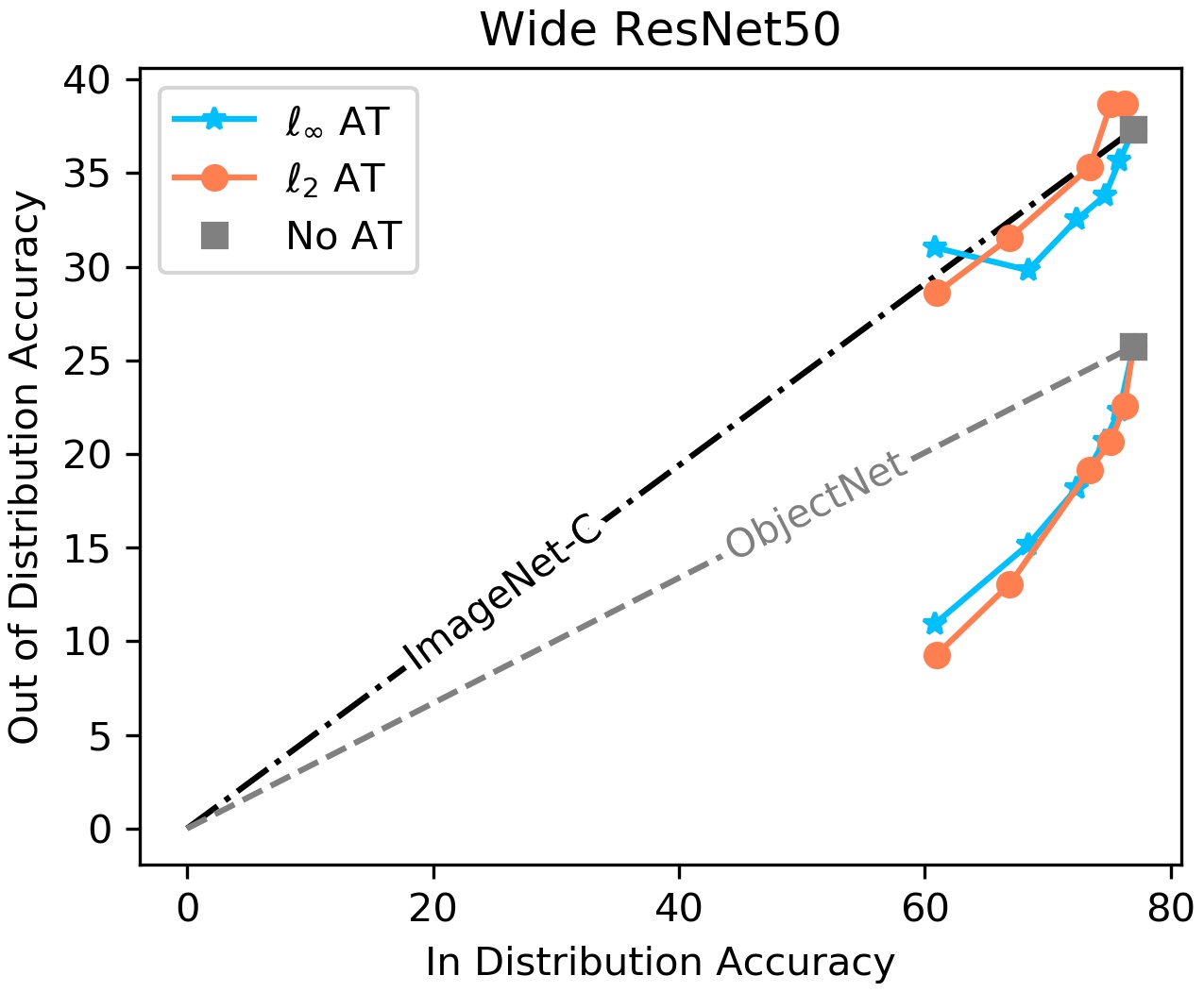}
\caption{ObjectNet, ImageNet-C, and ImageNet accuracies for WideResNet50s.}
\vspace{-0.5cm}
\label{fig:ood_wide}
\end{wrapfigure}

\noindent{{\bf WideResNets.}} We corroborate all our empirical findings on ResNet18s and ResNet50s on the WideResNet50 (2x) architecture. Figure \ref{fig:ood_wide} shows that accuracy drop in AT models is more severe on distribution shifts that break spurious correlations (ObjectNet), unlike the accuracy drop due to corruption of both core and spurious features (ImageNet-C), which can likely be explained by the reduced standard accuracy of AT models. 

\begin{figure}
    \centering
    \begin{minipage}{0.45\linewidth}
    \centering
    \includegraphics[width=0.9\textwidth]{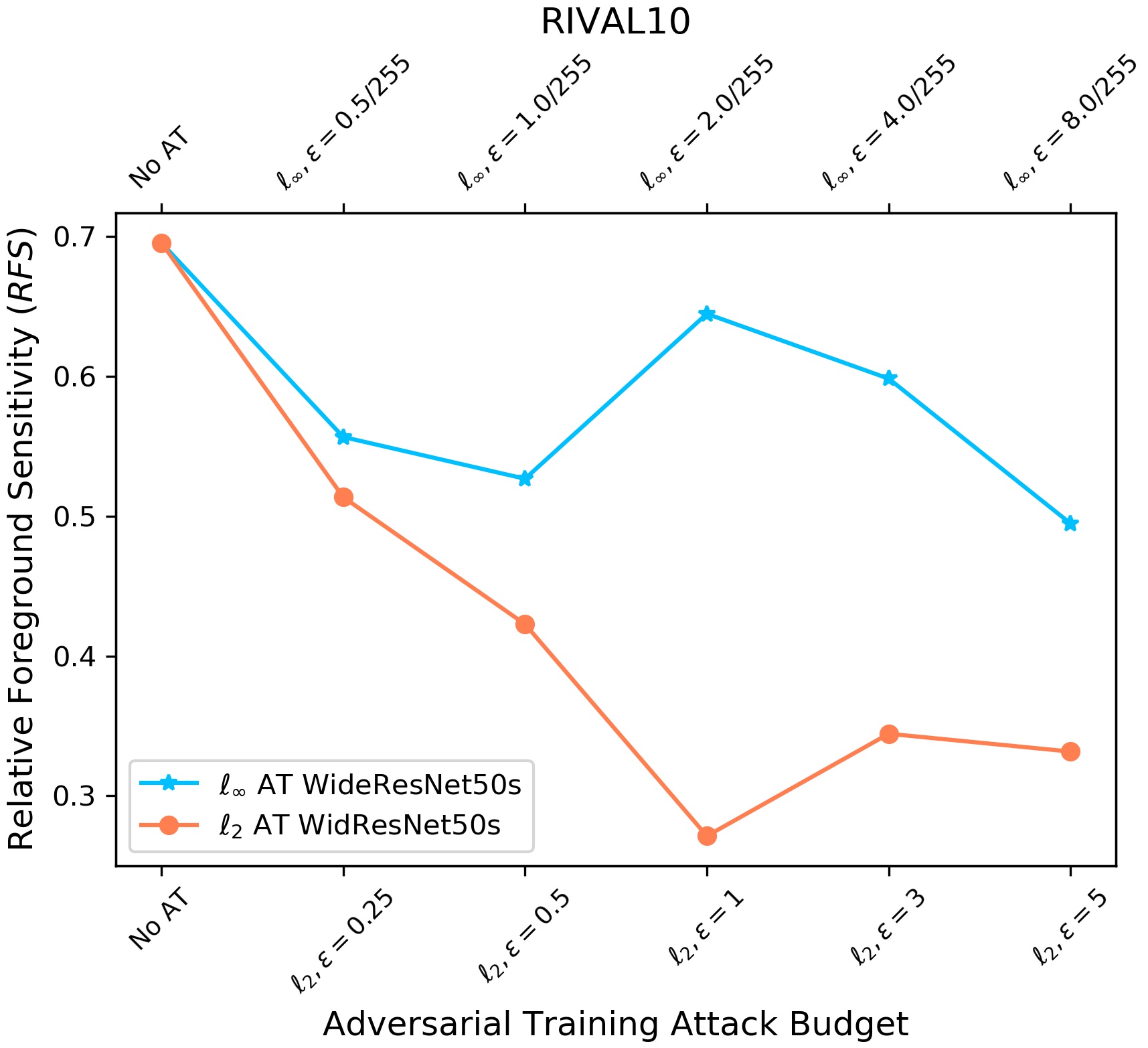}
    \end{minipage}
    \begin{minipage}{0.45\linewidth}
    \centering
    \includegraphics[width=0.9\textwidth]{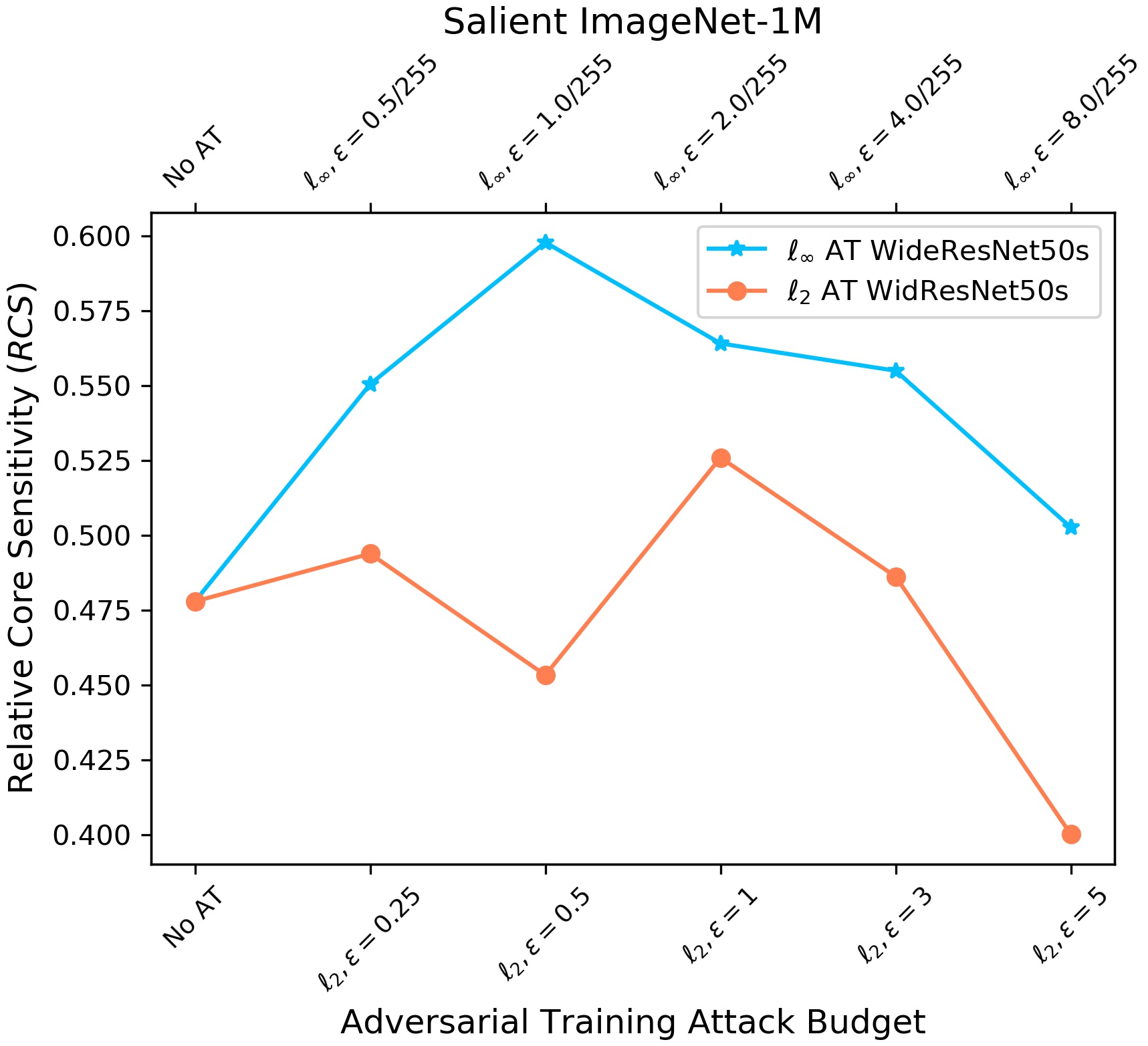}
    \end{minipage}
    \caption{$RFS$ and $RCS$ for WideResNet50s. Sensitivity to core and foreground regions are reduced for higher $\epsilon$, especially for $\ell_2$ AT models and for $RFS$, computed over the RIVAL10 dataset, where background correlations are stronger.}
    \label{fig:rfs_rcs_wide}
\end{figure}

\looseness=-1
Figure \ref{fig:rfs_rcs_wide} shows reduced sensitivity to core and foreground regions for AT models. Again, the effect is more pronounced for $\ell_2$ adversarially training and for larger $\epsilon$. Also, we again see that decrease in $RCS$ is less consistent than the drop in $RFS$. We conjecture that the diversity and fine-grain Salient ImageNet classification task reduces the strength of spurious correlations present in the data, thus diluting our observed effects of adversarial training on spurious feature reliance.

\begin{figure}
    \centering
    \begin{minipage}{0.45\linewidth}
    \centering
    \includegraphics[width=0.9\textwidth]{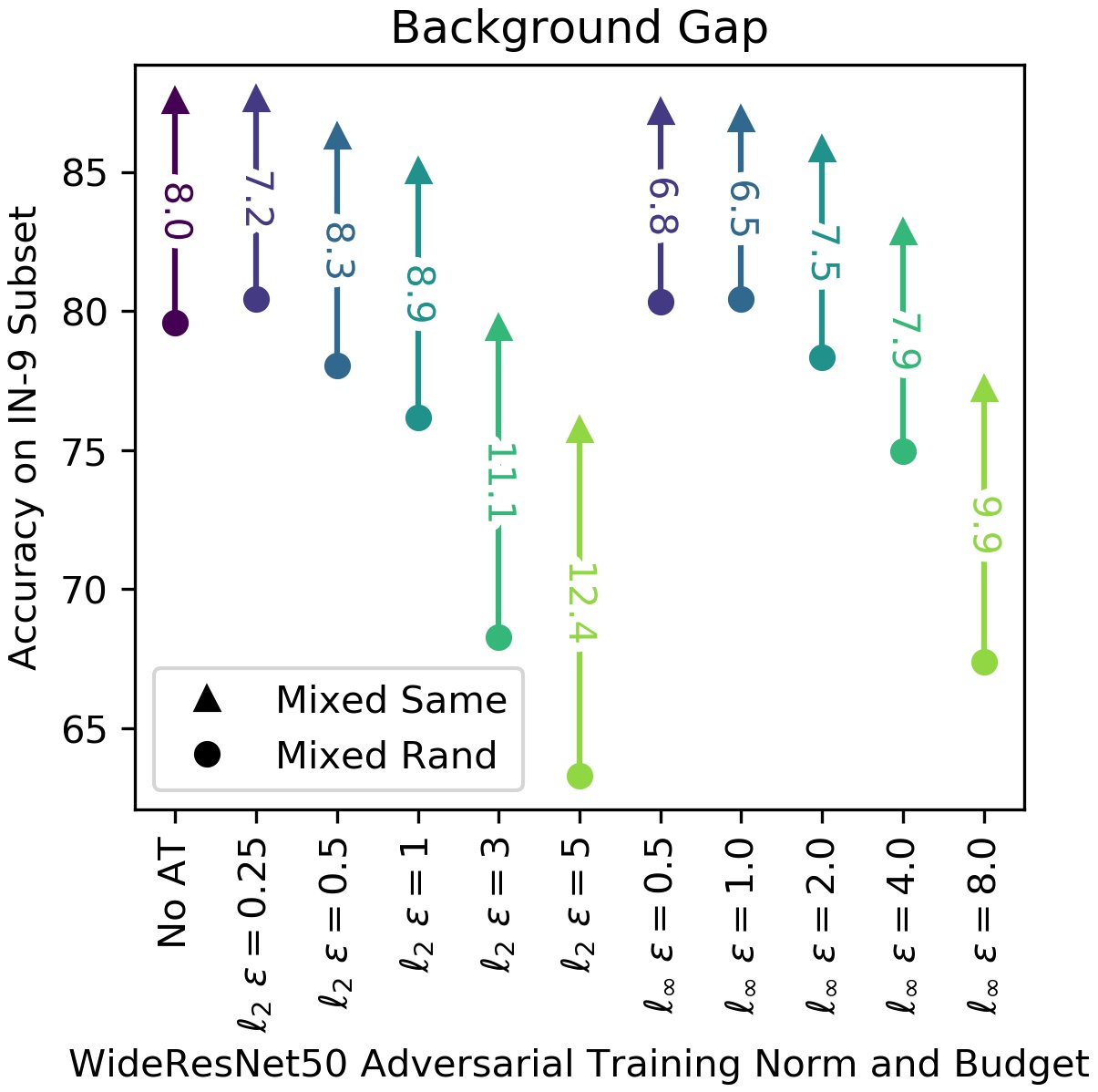}
    \end{minipage}
    \begin{minipage}{0.45\linewidth}
    \centering
    \includegraphics[width=0.9\textwidth]{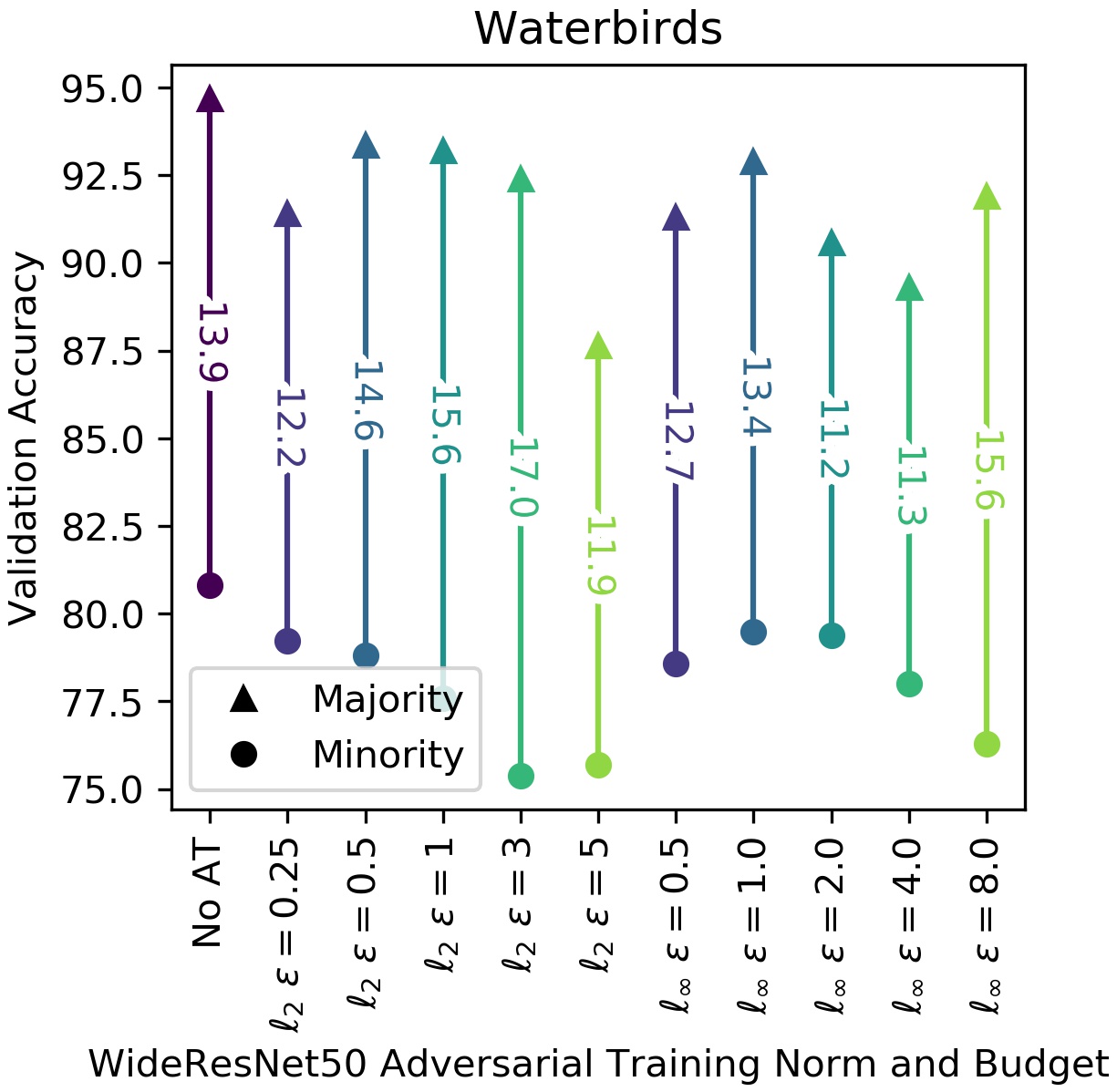}
    \end{minipage}
    \caption{Background Gap (IN-9) and Waterbirds gap for WideResNet50s. AT models, especially under $\ell_2$ norm, see larger accuracy drops when spurious correlations are broken.}
    \label{fig:in9_waterbirds_wide}
\end{figure}

Lastly, figure \ref{fig:in9_waterbirds_wide} shows the drop in accuracy due to breaking spurious background correlations is larger for AT models. Indeed, the absolute background gap (IN-9) for the WideResNet50 under $\ell_2$ AT with $\epsilon=5$ is $50\%$ larger than the gap for the standardly trained baseline. We note that the absolute gaps are smaller in some cases. We believe the lower standard accuracy of AT models may contribute to this, as there is less accuracy to drop from. Nonetheless, it is intriguing that in some cases, $\ell_\infty$ adversarial training seems to reduce spurious feature reliance; while our theory explains how a spurious feature can be completely ignored under $\ell_\infty$ training, it does not explain cases where spurious feature reliance is reduced compared to standard training. We believe this is an interesting direction for future work. 

\begin{figure}
    \centering
\begin{subfigure}[b]{0.98\textwidth}
\centering
  \includegraphics[width=0.3\linewidth]{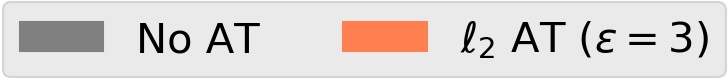}
  \caption{Legend. We compare $\ell_2$ adversarially trained models to standardly trained baselines for five new backbones.}
\end{subfigure}
\begin{subfigure}[b]{0.98\textwidth}
  \includegraphics[width=1\linewidth]{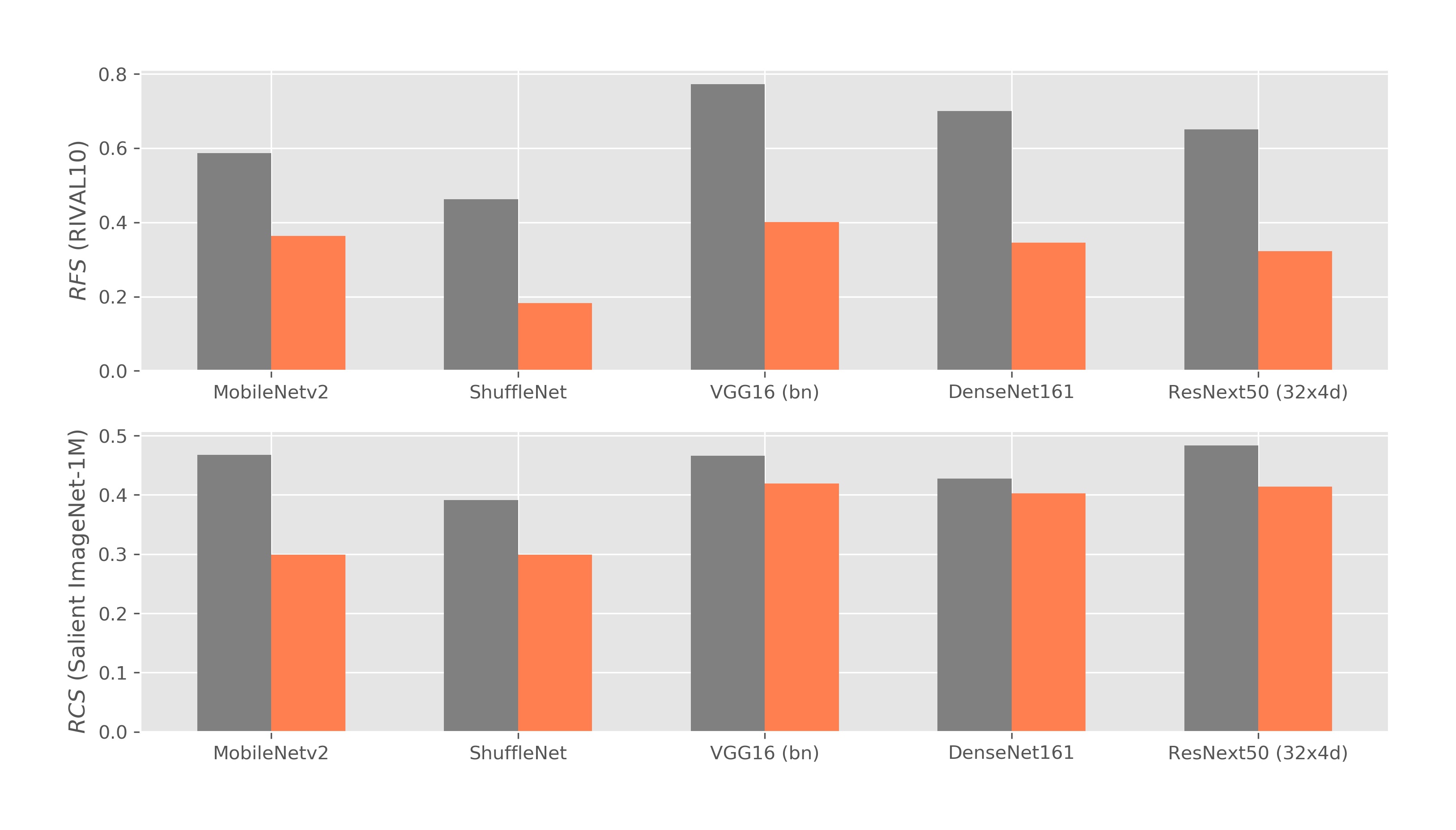}
  \caption{Lower $RFS$ ($RCS$) entails Lower Foreground (Core Feature) Sensitivity}
\end{subfigure}
\begin{subfigure}[b]{0.98\textwidth}
  \includegraphics[width=1\linewidth]{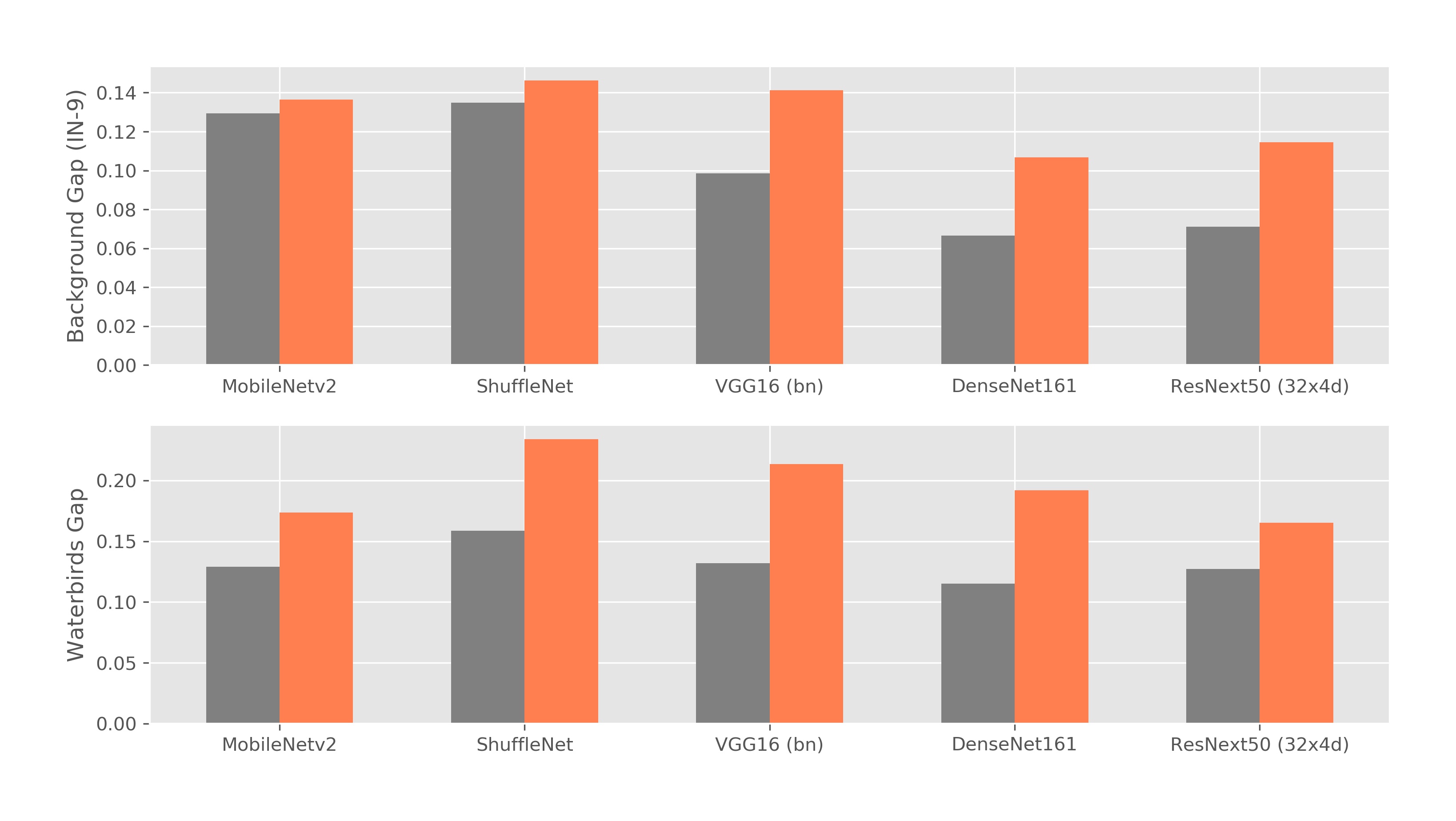}
  \caption{Higher Gap entails Greater Background/Spurious Sensitivity}
\end{subfigure}
\begin{subfigure}[b]{0.98\textwidth}
  \includegraphics[width=1\linewidth]{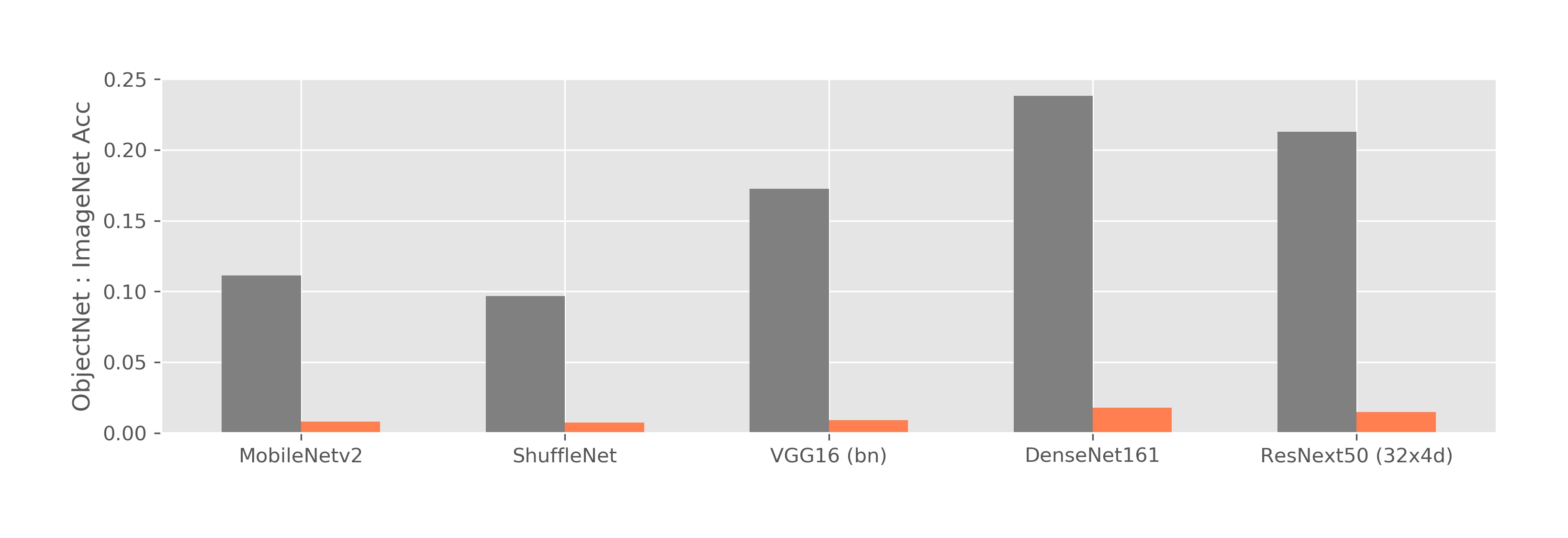}
  \caption{Lower Ratio entails Lower Natural Distributional Robustness}
\end{subfigure}
\caption{Corroborating findings on additional backbones.}
\label{fig:extra_backbones}
\end{figure}

\noindent{\bf{Other backbones.}} We now show results for ten other models, half of which are $\ell_2$ adversarially trained with $\epsilon=3$, while the others are standardly trained. Figure \ref{fig:extra_backbones} summarizes our results, corroborating each of our empirical findings. 